\documentclass{article}

\PassOptionsToPackage{numbers, compress}{natbib}
\usepackage[preprint]{neurips_2026}

\usepackage[utf8]{inputenc} 
\usepackage[T1]{fontenc}    
\usepackage{hyperref}       
\usepackage{url}            
\usepackage{booktabs}       
\usepackage{amsfonts}       
\usepackage{nicefrac}       
\usepackage{microtype}      
\usepackage{xcolor}         

\usepackage{amsmath,amsfonts,bm}

\def\eqref#1{equation~\ref{#1}}

\def\1{\bm{1}}

\DeclareMathAlphabet{\mathsfit}{\encodingdefault}{\sfdefault}{m}{sl}
\SetMathAlphabet{\mathsfit}{bold}{\encodingdefault}{\sfdefault}{bx}{n}

\usepackage[utf8]{inputenc} 
\usepackage[T1]{fontenc}    
\usepackage{hyperref}       
\usepackage{url}            
\usepackage{booktabs}       
\usepackage{amsfonts}       
\usepackage{nicefrac}       
\usepackage{microtype}      
\usepackage{xcolor}         
\usepackage{graphicx}
\usepackage{amsmath}
\usepackage{amssymb}
\usepackage{mathtools}
\usepackage{amsthm}
\usepackage[capitalize,noabbrev]{cleveref}
\usepackage{ml_defs}
\usepackage{threeparttable}
\usepackage{fontawesome5}
\usepackage{subcaption}
\usepackage[inline]{enumitem}
\usepackage{adjustbox}
\usepackage{float}
\usepackage{bbm}
\usepackage{titletoc}

\usepackage{wrapfig}
\usepackage{tcolorbox}
\usepackage{siunitx}
\usepackage{multirow}
\usepackage{algorithm}

\newcommand{\ourmethod}{Bi-FORK}

\renewcommand{\paragraph}[1]{\textbf{#1}. }

\title{Bi-FORK: Generative Modeling of High-Dimensional Bifurcating Systems}

\author{Anna Zimmel$^{*\,1}$ Fleur Hendriks$^{*\,2\,3\,4}$ Markus Holzleitner$^{1}$ Florian Sestak$^{1}$ \\ 
\textbf{Martin Weichselbaumer}$^{1}$ \textbf{Vlado Menkovski}$^{\dagger \, 4}$ \textbf{Johannes Brandstetter}$^{\dagger \,1\,5}$ \\
$^{*}$ Equal contribution as first authors, $^{\dagger}$ Equal contribution as last authors \\
$^{1}$ELLIS Unit, LIT AI Lab, Institute for Machine Learning, JKU Linz \\
$^{2}$Department of Mechanical Engineering, Eindhoven University of Technology\\
$^{3}$DIFFER – Dutch Institute for Fundamental Energy Research\\
$^{4}$Department of Mathematics and Computer Science, Eindhoven University of Technology\\
$^{5}$Mistral, Paris \\
\texttt{zimmel@ml.jku.at, f.hendriks@differ.nl} \\
}

\begin{document}

\maketitle

\begin{abstract}
Bifurcations are ubiquitous in physical systems, from structural buckling to fluid and climate dynamics, yet they remain largely unexplored in deep learning. At a symmetry-breaking bifurcation, a single input admits multiple equally valid solutions, violating the one-to-one assumption underlying most learned physical surrogates. We introduce Bi-FORK, a generative framework for learning these one-to-many solution maps in high-dimensional systems. Bi-FORK generates complete trajectories through latent flow matching, preserving space and time coherence, and uses repulsion-guided sampling to recover distinct solution branches in a single amortized pass. We evaluate Bi-FORK on buckling beams, mechanical metamaterials, and Allen-Cahn phase separation, spanning continuous, discrete, and field-valued bifurcations with discretizations up to 260,000 points. Bi-FORK recovers the multimodal solution structure while scaling several orders of magnitude beyond prior approaches, opening generative modeling to high-dimensional bifurcating physical systems.
\end{abstract}

\section{Introduction} \label{sec:introduction}
Bifurcations are fundamental to nonlinear physical systems and occur across structural mechanics, fluid dynamics, climate dynamics, and phase-transition phenomena \citep{ruelle1971nature, steffen2018trajectories, rietkerk2021evasion}.
A bifurcation is a qualitative change in the number or stability of solutions as a control parameter varies.
When a bifurcation breaks a symmetry, it produces a one-to-many mapping from control parameter to system response: multiple, equally valid solutions rather than one.
For example, a compressed column is stable up to a critical load, then bifurcates into symmetric buckled branches~\citep{bavzant2003stability}.
Classical numerical methods, including finite-element and finite-volume methods, can compute these solutions accurately.
However, for high-dimensional systems, recovering the full bifurcation structure requires stability and mode analyses together with repeated continuation or deflation solves \citep{brow1971deflation, farrell2015deflation}, which is computationally expensive.
Therefore, machine learning (ML) methods are an attractive alternative as they amortize the cost of building the training set over the many fast evaluations of unseen system conditions \citep{abramson2024accurate, alkin2024universal, alkin2025abupt, giacomini2026surrogates, tang2026learning}.


However, because the system is one-to-many, training discriminative ML models averages over the solution space and, as a consequence, produces nonphysical predictions \citep{rupprecht2017learning, kochkov2024neural,price2023gencast}.
This has motivated the development of probabilistic methods based on GANs, diffusion models, and more recently flow matching, typically framed around uncertainty or the loss of fine-scale structure \citep{gao2023prediff,yoon2023deterministic,calzolari2025accelerating}.
Most probabilistic physical surrogates model epistemic or unresolved uncertainty; symmetry-breaking bifurcations instead produce genuinely multiple, equally valid solutions structured by the system's symmetry.
Only a few recent approaches treat the bifurcation structure directly as the cause of solution multiplicity \citep{fabiani2021numerical,pichi2023artificial,zou2025learning, hendriks2026equivariantflowmatch,guilhoto2026multimodal, romero2026note}.
Nevertheless, these approaches recover systems of at most a few hundred degrees of freedom with small, known, discrete symmetry groups.
\begin{figure}[t]
    \centering
    \includegraphics[width=\linewidth]{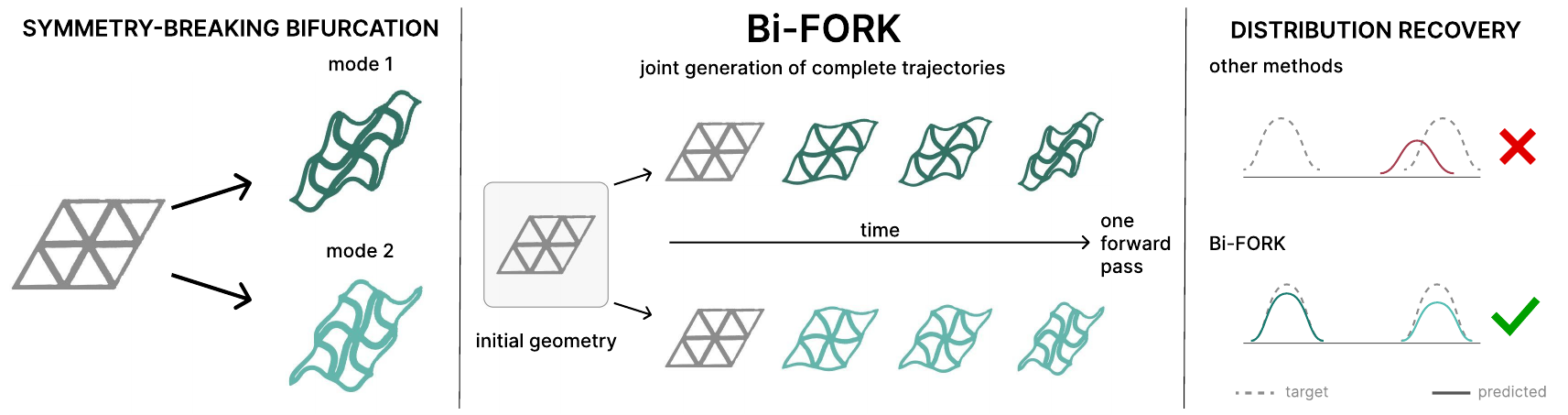}
    \caption{Learning bifurcating solution distributions. Left: Symmetry breaking results in multiple valid trajectories for the same control parameter and initial state. Center: Latent flow matching jointly samples trajectories and resolves distinct branches in one forward pass. Right: Unlike mode-averaging baselines, \ourmethod~recovers the target distribution.}
    \label{fig:figure1}
\end{figure}

\paragraph{The main challenge} Scaling to larger systems is challenging because recovering a bifurcating distribution requires every part of the system to \emph{agree} on which branch it is realizing in both space \emph{and} time.
A beam buckling to the left at one end and right at the other is an invalid solution, as is a beam that first buckles to the left and then suddenly switches to buckling to the right halfway through a trajectory.
Moreover, bifurcations are typically rare events in trajectories.
Training a frame-by-frame autoregressive model to produce trajectories would face significant challenge of capturing this key aspect of the system's behavior while being underrepresented in the training data.
Predicting the full trajectory allows the model to instead represent the branch choice globally in latent space, for all steps and elements at the same time.
Mesh-based GNN surrogates communicate through local message passing, making such global coordination increasingly difficult as systems grow due to oversmoothing and oversquashing \citep{oono2019graph, chen2020measuring, cai2020note, alon2021bottleneck}.

Evaluation is similarly nontrivial.
Pointwise metrics such as rank histograms and CRPS \citep{gronquist2021deep,gao2023prediff,he2025skillful,mahesh2024huge,larsson2025crps} assess pointwise calibration but cannot determine whether samples are coherent trajectories, while best-of-$K$ scores ignore most of the predicted distribution.
Generic two-sample distances can detect mode collapse but become difficult to estimate in high dimensions and do not exploit the symmetry structure of bifurcating solutions \citep{hamill2001interpretation,bischoff2024practical}.

We address these challenges with \ourmethod: Bifurcation Flow-matching Over Repulsion-guided Kinematics, a latent flow-matching framework for symmetry-breaking bifurcations (see \cref{fig:figure1}).
\ourmethod~generates full trajectories jointly, keeping branch choices consistent across space and time rather than learning them through local transitions.
Because full-resolution execution is intractable, flow matching operates in a compact latent space, decoupling global coordination from discretization.
Although compression can average modes, branch identity dominates trajectory variance and is therefore preserved by the reconstruction objective without supervision.
At inference, particle guidance \citep{unterthiner2017coulomb, corso2024particle} repels concurrent samples, acting as a parallelized analog of classical deflation \citep{brow1971deflation,farrell2015deflation} to recover distinct solutions.
We assess the resulting distribution by matching predictions to their nearest ground truth and, per input, measuring orbit coverage under the relevant symmetry group via the Jensen-Shannon divergence, thereby detecting regime collapse and quantifying coverage of the underlying distribution.

We evaluate the approach on three symmetry-breaking bifurcation problems: buckling beams, buckling mechanical metamaterials, and phase separation described by Allen-Cahn. These systems cover two- and three-dimensional grid- and mesh-based problems, various symmetry groups (both discrete and continuous), trajectories of 32 to 200 timesteps, and discretizations from a handful of points to 260{,}000.
To our knowledge, this is the first demonstration of solution distribution recovery for one-to-many problems at that scale, and we show that our method outperforms existing approaches in both accuracy and efficiency.

Our contributions are:
\begin{enumerate}
    \item A latent flow-matching framework that models whole trajectories, making branch coherence structural, and scales several orders of magnitude beyond prior work.
    \item A particle-guidance sampling mechanism recovering multiple distinct solutions in one amortized pass, as a parallel analog of deflation.
    \item Diagnostics based on symmetry-orbit matching and per-input Jensen-Shannon divergence that detect regime collapse and extend distributional verification to continuous and infinite symmetry groups.
\end{enumerate}

\section{A Blueprint for Bifurcating Regimes}
\label{sec:bifurcation_learning_problem}

\subsection{Bifurcation Systems}\label{subsec:bifurcating}
\begin{wrapfigure}{r}{0.35\textwidth}
\centering
    \vspace{-1.5em}
    \includegraphics[width=\linewidth]{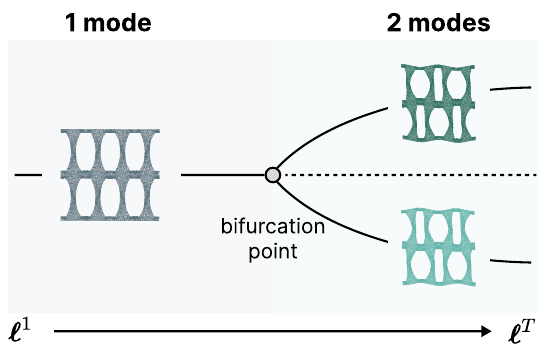}
    \caption{Exemplary pitchfork bifurcation for the buckling metamaterials use case with increasing control protocol $\boldsymbol{\ell}$
    }
    \vspace{-1.8em}
    \label{fig:illustrative_bifurcation}
\end{wrapfigure}
Recall that a \emph{bifurcation} is the loss of uniqueness of a stable equilibrium as a control parameter crosses a critical value, with buckling as the canonical mechanical example~\citep{koiter1945, antman2005,hendriks2026equivariantflowmatch}.

\paragraph{Setting} We represent trajectories as either point clouds or scalar fields on a regular grid. In both cases, $\bfM^t\in\dR^{N\times C}$ denotes the modeled response at $N$ spatial locations and timestep $t\in\{1,\dots,T\}$, and $\bfM^{1:T}\in\dR^{T\times N\times C}$ forms the complete response trajectory.
For point clouds, the $N$ locations are material points with reference coordinates $\bfX^1\in\dR^{N\times D}$, where $D\in\{2,3\}$. The response $\bfM^t$ is their displacement or microfluctuation field, so $C=D$; deformed coordinates $\bfX^t$ are recovered from $\bfX^1$ and $\bfM^t$ using the prescribed loading and dataset-specific kinematics.
For regular-grid data, the $N$ locations are fixed grid sites and $\bfM^t$ is the scalar field itself, so $C=1$.



\paragraph{Datasets} We consider three physical systems: (i) \textit{Beam3D} models a slender beam that is compressed from one end until it buckles in arbitrary directions. (ii) \textit{Mechanical Metamaterials} captures the buckling behavior of 2D porous materials with different periodic patterns, leading to distinct stable shapes. (iii) \textit{Allen-Cahn} describes how a mixture separates into different phases over time, with a shared homogeneous initial state that evolves into several possible stable configurations.

\paragraph{Modes as a symmetry orbit}
A problem instance is a \emph{condition} $\mathbf{c} = (\bfX^1, \boldsymbol{\ell})$: a reference geometry together with a control protocol $\boldsymbol{\ell} = (\boldsymbol{\ell}^1, \dots, \boldsymbol{\ell}^T)$ that is fixed in advance and not influenced by the system's response. The set of all feasible conditions we denote by $\mathcal{C}$.
For the mechanical metamaterials datasets, for example, $\boldsymbol{\ell}$ is a quasi-static loading schedule generated by a single terminal load ($\bfF^t$ interpolating toward $\bfF^{T}$), and the index $t$ enumerates states and load increments simultaneously: $\bfM^t$ is the equilibrium microfluctuation response to the input $\boldsymbol{\ell}^t$ (see \cref{fig:illustrative_bifurcation}).

For Allen-Cahn, the protocol is constant, $\boldsymbol{\ell}^t \equiv (\epsilon, \mu)$, where $\epsilon$ is a length scale and $\mu$ determines local phase structure.
The systems are \emph{multistable}: under a fixed condition, several physically distinct outcomes---\emph{modes}---are admissible. We denote their mode set by $\mathcal{M}_{\mathbf{c}}$\footnote{Here `Mode' means a distinct admissible outcome, \emph{not} an eigenmode: two symmetry-related metamaterial solutions stem from one buckling eigenmode but count as two modes. The usage agrees with statistics---outcomes follow $p(\cdot \mid \mathbf{c})$, and for finite mode sets each branch is an isolated local maximum of this distribution; for the beam the distribution is uniform along the orbit, which is then its maximizer set.}, and write $\mathcal{M}_{\mathbf{c}}^t$ for the set of their responses at timestep $t$.
Symmetry organizes this set: with
\begin{align}
G_\mathbf{c} = \{\, g \mid g \cdot \mathbf{c} =  \mathbf{c} \  \,\}
\end{align}
the symmetry group of the condition, any response $\bfM$ yields further solutions $g\cdot\bfM$, so the mode set contains full orbits,
\begin{align}
    G_\mathbf{c} \cdot \bfM_\mathbf{c}^t \subseteq \mathcal{M}_{\mathbf{c}}^t,
    \qquad
    H_\mathbf{c}^t = \{\, g \in G_\mathbf{c} \mid g \cdot \bfM_\mathbf{c}^t = \bfM_\mathbf{c}^t \,\},
\end{align}
where $\bfM_\mathbf{c}^t$ is a representative response and $H_\mathbf{c}^t$ its isotropy subgroup; each orbit is isomorphic to $G_\mathbf{c}/H_\mathbf{c}^t$ with size $[G_\mathbf{c} : H_\mathbf{c}^t]$.
For Beam3D and the mechanical metamaterials, the inclusion is an equality (a single orbit); for Allen-Cahn, the mode set is a union of orbits, one per morphology.
\Cref{tab:symmetry_structure} instantiates $G_\mathbf{c}$, $H_\mathbf{c}$, and $\mathcal{M}_{\mathbf{c}}$ per dataset; note e.g.\ the beam, where $G_\mathbf{c} = \mathrm{O}(2)$ and the buckled state retains one reflection, giving the circular orbit $\mathrm{O}(2)/\mathbb{Z}_2 \cong \mathrm{SO}(2)$.

\begin{table}[t]
    \centering
    \caption{Symmetry structure per dataset: group $G_\mathbf{c}$ of the condition, isotropy $H_\mathbf{c}$ of an output state, resulting mode set $\mathcal{M}_{\mathbf{c}} \cong G_\mathbf{c}/H_\mathbf{c}$ (union of orbits for Allen-Cahn).}
    \label{tab:symmetry_structure}
    \footnotesize
    \begin{tabular}{@{}lllll@{}}
        \toprule
        \textbf{Dataset} & $G_\mathbf{c}$ & $H_\mathbf{c}$ & $\mathcal{M}_{\mathbf{c}}$ & $|\mathcal{M}_{\mathbf{c}}|$ \\
        \midrule
        Beam3D & $\mathrm{O}(2)$ & $\mathbb{Z}_2$ & circle of directions $\phi$ & $\infty$ \\
        Metamaterials & wallpaper subgroup fixing $\boldsymbol{\ell}$ & pattern stabilizer & translated patterns & $1, 2, 4$ \\
        Allen-Cahn & domain syms $\times \{u \mapsto \pm u\}$ & morphology symmetries & orbits of morphologies & varies \\
        \bottomrule
    \end{tabular}
\end{table}

\paragraph{Bifurcation time}
We define the bifurcation time as the first step at which the solution loses symmetry,
\begin{align}
t_b = \min \{t\in\{2,\dots,T\} \mid H_\mathbf{c}^t \subsetneq H_\mathbf{c}^{t-1} \}.
\end{align}
In practice, we estimate $t_b$ as the first step at which trajectories under the same condition separate beyond a small threshold.
This also accommodates Allen-Cahn, where all trajectories share the same homogeneous initial state but macroscopic modes emerge only later. The choice of representation, the practical estimator, and secondary bifurcations are discussed in Appendix \ref{app:bifurctime_context}.
Probabilistically, $t_b$ marks the onset of distinct branches in the distribution of outcomes conditioned on the context $\bf c$ at timestep $t$,
connecting physical symmetry breaking to the need for distribution-valued prediction.

\paragraph{Three regimes of multimodality}
Our datasets span discrete ($|\mathcal{M}_{\mathbf{c}}| \in \{1,2,4\}$, mechanical metamaterials of \citealp{hendriks2025wallpaper}), continuous ($|\mathcal{M}_{\mathbf{c}}| = \infty$, Beam3D), and field-valued (Allen-Cahn) mode sets; see Appendix \ref{app:datasets}.

\subsection{Learning task}
\label{subsec:general_learning_task}

Given a condition $\mathbf{c}$, the task is to learn the conditional distribution over all feasible trajectories
\begin{equation}
    p\!\left(\bfM^{1:T}\mid\mathbf{c}\right),
\end{equation}
where the support of the target distribution is the mode set $\mathcal{M}_{\mathbf{c}}$.
The training datasets differ in their coverage of this support: the mechanical metamaterials enumerate all modes per condition, whereas Beam3D and Allen-Cahn provide samples from the mode set.
We assume unknown perturbations that push the system toward one solution within an orbit have no preferred direction. Under this assumption, the condition's symmetry implies that modes within an orbit are equally likely, so for single-orbit datasets the target is uniform over $\mathcal{M}_{\mathbf{c}}$.
Unlike standard trajectory prediction, the mapping from $\mathbf{c}$ to the system evolution is one-to-many: multiple physically valid trajectories arise under the same condition.
During inference, the model receives $\mathbf{c}$ and generates one or more trajectory samples from a learned distribution $p_{\theta_{\mathrm{flow}}}(\bfM^{1:T}\mid\mathbf{c})$.
\subsection{Modeling}
\label{subsec:modeling_bf}

\begin{figure}[H]
    \centering
    \includegraphics[width=\linewidth]{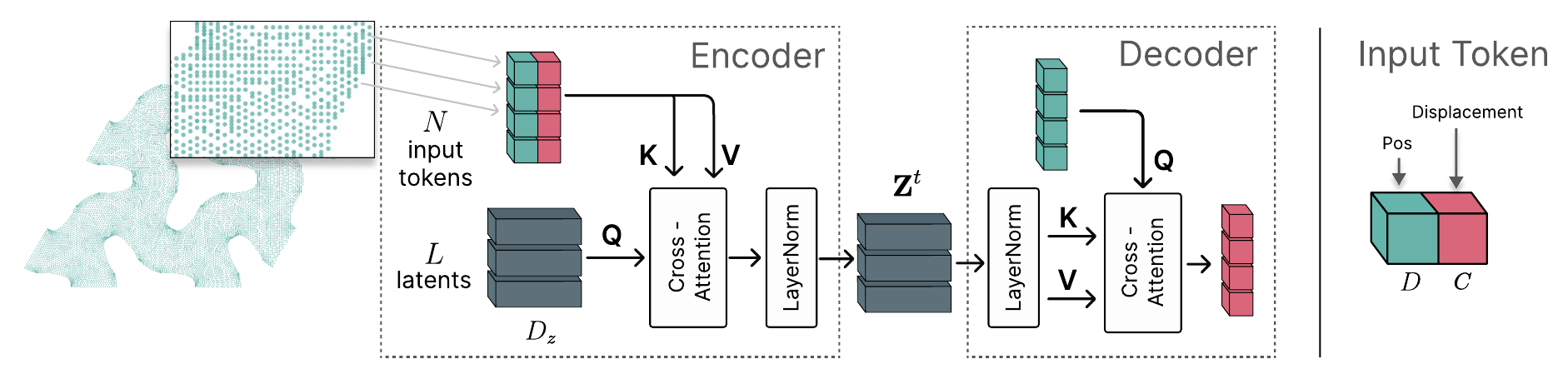}
    \caption{Encoder-decoder structure for point cloud datasets. Input tokens (reference positions $\bfX^1$ and displacements) are encoded into latent representation by cross-attending to $L$ learned latent query tokens. The decoder reconstructs the displacements conditioned on the reference positions.}
    \label{fig:autoencoder_architecture}
\end{figure}

\paragraph{Autoencoder} We adopt a Perceiver-style cross-attention architecture~\citep{jaegle2021perceiver,alkin2024universal,sestak2026lamslide} for all point cloud datasets.
For each timestep $t$, the encoder maps the reference configuration $\bfX^1$ and response field $\bfM^t$ to $L$ latent tokens of dimension $D_z$, i.e., $\bfZ^t\in\mathbb{R}^{L\times D_z}$, using cross-attention with learned latent queries, followed by self-attention on the latent set.
Importantly, the decoder reconstructs $\bfM^t$ by querying these tokens at the fixed \emph{reference} positions $\bfX^1$.
Thus, the latent representation $\bfZ^t$ encodes all timestep-dependent states, while spatial reconstruction is always anchored to the same reference geometry $\bfX^1$.
This enables the independent analysis and prediction of latent trajectories $\{\bfZ^t\}_{t=1}^T$, which can then be decoded at the reference positions to recover $\bfM^t$.
For regular grid datasets, we use a 3D Vision Transformer (ViT) \citep{dosovitskiy2020image, prabhakar2024vit}, detailed in the respective section (\cref{subsec:allencahn_exp}).

\paragraph{Latent flow matching}
For the approximator, we draw inspiration from the flow-based model framework in~\cite{sestak2026lamslide}.
Given $T_{\mathrm{obs}}$ known states $\bfZ^{1:T_{\mathrm{obs}}}$ and the condition $\mathbf{c}$, it generates the remaining states $\bfZ^{T_{\mathrm{obs}}+1:T}$.
The flow-based model then learns a path between noise and data.
Let $\bfZ_1 := \bfZ^{1:T} \in \dR^{T \times L \times D_z}$ be the true latent trajectory and $\boldsymbol{\xi}\sim\mathcal{N}(0,I)$ be noise. For every $\tau\in[0,1]$, a probability path is specified by
\begin{align}
    \bfZ_\tau = \tau\,\bfZ_1 + (1-\tau)\,\boldsymbol{\xi},
    \label{eq:interpolant}
\end{align}
such that pure noise is obtained with $\tau=0$, and the true trajectory at $\tau=1$. We train a network with parameters $\theta_{\mathrm{flow}}$ to reverse the applied corruption: given the noised trajectory, the flow time $\tau$, observed states and the control protocol, the clean trajectory is predicted by
\begin{align}
    f_{\theta_{\mathrm{flow}}} : (\bfZ_\tau,\tau,\bfZ^{1:T_{\mathrm{obs}}},\mathbf{c}) \mapsto \widehat{\bfZ}_1 \in \dR^{T \times L \times D_z}.
    \label{eq:denoiser}
\end{align}

We adapt the standard flow matching approach through optimal transport-based coupling of in-batch pairs to minimize the path length between the noise and data distributions~\citep{tong2023improving}. Additionally, we require the model to predict the bifurcation point, represented by a binary mask $\widehat{\bfb} \in \{0,1\}^{T}$. At inference, $\widehat{\bfZ}_1$ is recovered by solving the probability-flow ODE associated with Eq.~\ref{eq:interpolant}, starting from pure noise.

All timestep-dependent information for the prediction is expressed by $\bfZ_\tau$ and its conditioning signals. The formulation is model agnostic to the specific network architecture and training procedure used for $f_{\theta_{\mathrm{flow}}}$ (see Appendix~\ref{app:architecture} for details).

\paragraph{Inference using particle guidance}
Due to the \emph{multimodal} nature of the problem, the control protocol and initial state do not determine which mode the system settles into, e.g., the orbit coordinate (e.g.\ the buckling direction $\phi$) is left free by $\mathbf{c}$.
While the generative model predicts a single valid mode rather than averaging across them, sampling alone does not control the distribution over modes.
\begin{wrapfigure}{r}{0.42\textwidth}
\centering
    \includegraphics[width=\linewidth]{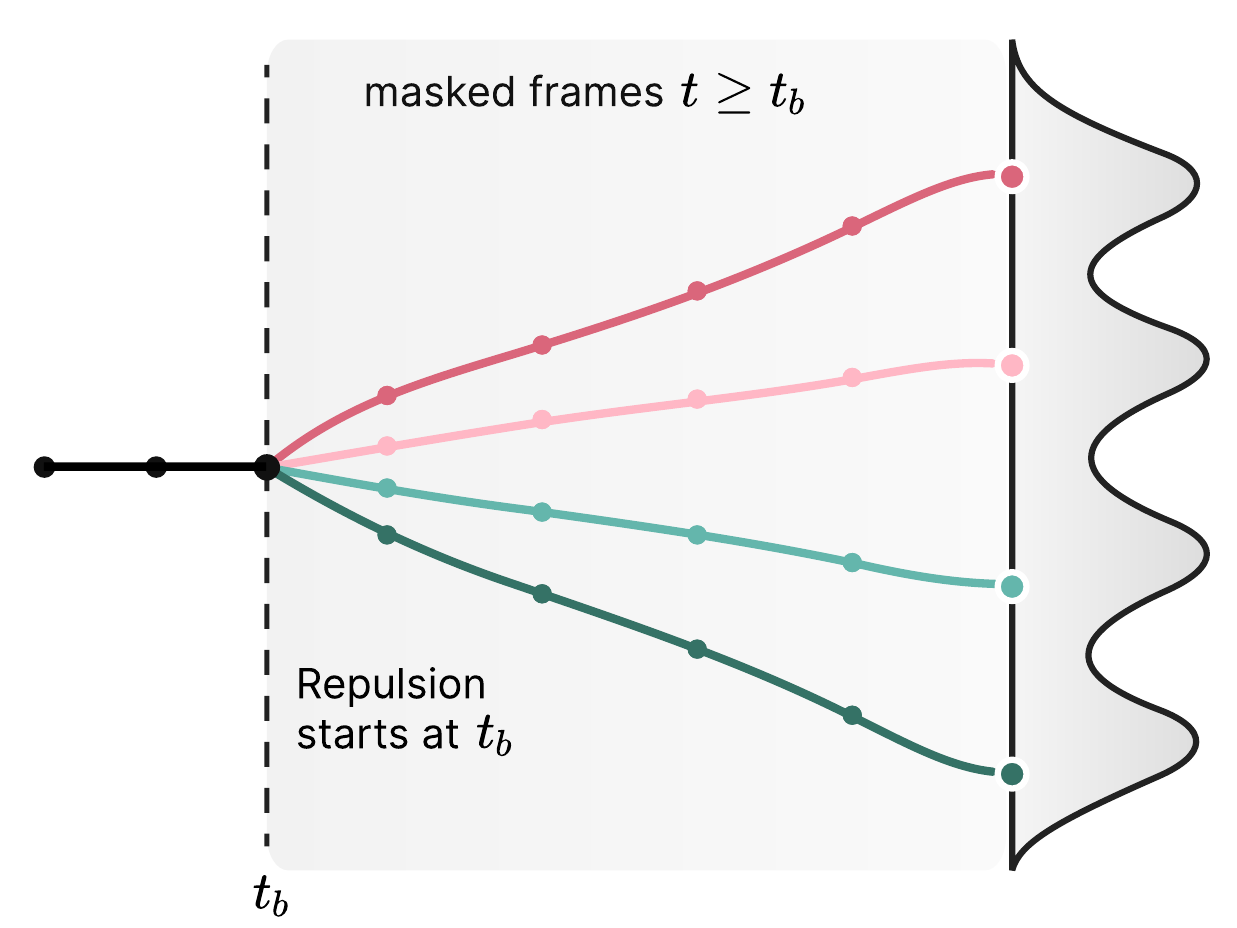}
    \caption{Particle guidance during sampling. Trajectories evolve identically up to bifurcation point $t_b$. From $t_b$, the repulsion term acts on the masked frames $t \geq t_b$, pushing the $K$ trajectories apart, yielding a multimodal distribution.}
    \label{fig:pg_approach_method}
    \vspace{-2.3em}
\end{wrapfigure}

We address this at inference with an approach inspired by particle guidance~\citep{corso2024particle}.
Rather than drawing $K$ trajectories independently, we sample $K$ candidate trajectories $\bfZ_\tau^{(1)}, \dots, \bfZ_\tau^{(K)}$ in a shared batch and integrate the probability-flow ODE for all of them jointly, leaving $f_{\theta_{\mathrm{flow}}}$ unmodified from training.
To maintain diversity, we add an inter-sample repulsion term to the velocity field at each step (\cref{fig:pg_approach_method}).

Each $\widehat{\bfZ}_1^{(k)}$ and its ordinary velocity field are computed independently via Eq.~\ref{eq:denoiser}, as at training time.
We additionally compute a repulsive term restricted to bifurcation-relevant frames, elected by the shared predicted mask $\widehat{\bfb}\in\{0,1\}^{T}$, where $\widehat{b}^t=1$ indicates a predicted post-bifurcation frame.
For each trajectory, the predicted values at these frames are masked and flattened into a single point; pairwise distances between trajectories' points are weighted by a Gaussian RBF kernel with bandwidth set to the median pairwise distance.
Each trajectory's displacement $\mathrm{force}^{(k)}$ is the resulting kernel-weighted average of unit vectors pointing away from its neighbors, so trajectories that are similar at the bifurcation frames are pushed apart while already-distinct trajectories barely interact. The repulsive displacement modifies the velocity field
$\widehat{v}_{\theta_{\mathrm{flow}}}=(\widehat{\bfZ}_1-\bfZ_\tau)/(1-\tau)$
of the probability-flow ODE for sample $k$.
Schematically,
\begin{align}
\widehat{v}^{(k)}
= w_{\mathrm{vel}}\,
\widehat{v}_{\theta_{\mathrm{flow}}}\bigl(\bfZ_\tau^{(k)}, \tau\bigr)
+ w_{\mathrm{rep}}\,
\widehat{\bfb} \odot \mathrm{force}^{(k)},
\label{eq:guided-velocity}
\end{align}
where $w_{\mathrm{vel}}$ weights the model velocity, $w_{\mathrm{rep}}$ controls repulsion strength, and $\widehat{\bfb}$ selects the predicted post-bifurcation frames.
The mask is broadcast over latent tokens and channels.
This simplified expression omits the time-dependent scaling and normalization which we describe in Appendix~\ref{app:guidance}.

As a result, all $K$ trajectories are integrated jointly rather than independently: at likely branch points, samples that would otherwise collapse onto the same mode are pushed apart, so the sample set spreads across distinct branches instead of concentrating on one.





\subsection{Metrics}
\label{subsec:metrics}
Evaluating generative models of bifurcating dynamical systems requires metrics that jointly capture trajectory-level precision and the distributional fidelity of predicted mode probabilities.
Standard metrics such as MSE are inadequate in this setting, as they neither reflect the multimodal structure of the output space nor penalize mode collapse.
We adopt two complementary metrics, adapted to systems with $|\mathcal{M}_{\mathbf{c}}| \in \{1, 2, 4\}$ or $|\mathcal{M}_{\mathbf{c}}| = \infty$ responses depending on the condition $\mathbf{c}$.

\textit{Mode Conditioned MAE (MCon-MAE)} quantifies trajectory-level precision. For each condition $\mathbf{c}$, we draw $K$ samples from the model and assign each to its nearest ground-truth mode by Euclidean distance. MCon-MAE is the resulting reconstruction error, averaged over samples and conditions:
\begin{equation}
    \label{eq:mcon_mae}
    \text{MCon-MAE} = \frac{1}{|\mathcal{C}|} \sum_{\mathbf{c} \in \mathcal{C}} \frac{1}{K}
\sum_{k=1}^{K} \operatorname{mean}\!\left(\left|\widehat{\bfX}_{\mathbf{c}}^{(k)} - \bfX_{\mathbf{c},m_{\star}^{(k)}}\right|\right)
\end{equation}
with mode assignment
$m_{\star}^{(k)} = \arg\min_{m \in \mathcal{M}_{\mathbf{c}}}
\left\| \widehat{\bfX}_{\mathbf{c}}^{(k)} - \bfX_{\mathbf{c},m} \right\|_F^2$, 
where $\bfX_{\mathbf{c},m}$ highlights the dependence on the condition $\mathbf c$ and mode $m$.
We further report the \textit{Jensen-Shannon divergence (JSD)} between the predicted and ground-truth mode distributions, estimated from $K$ samples drawn at inference for each condition.
This measures the extent to which a model represents each outcome with the correct relative frequency, rather than favoring a subset of modes.

\section{Experiments}
\label{sec:experiments}

\subsection{Beam3D} \label{subsec:beam_exp}
\begin{figure}[ht]
    \centering
    \includegraphics[width=\linewidth]{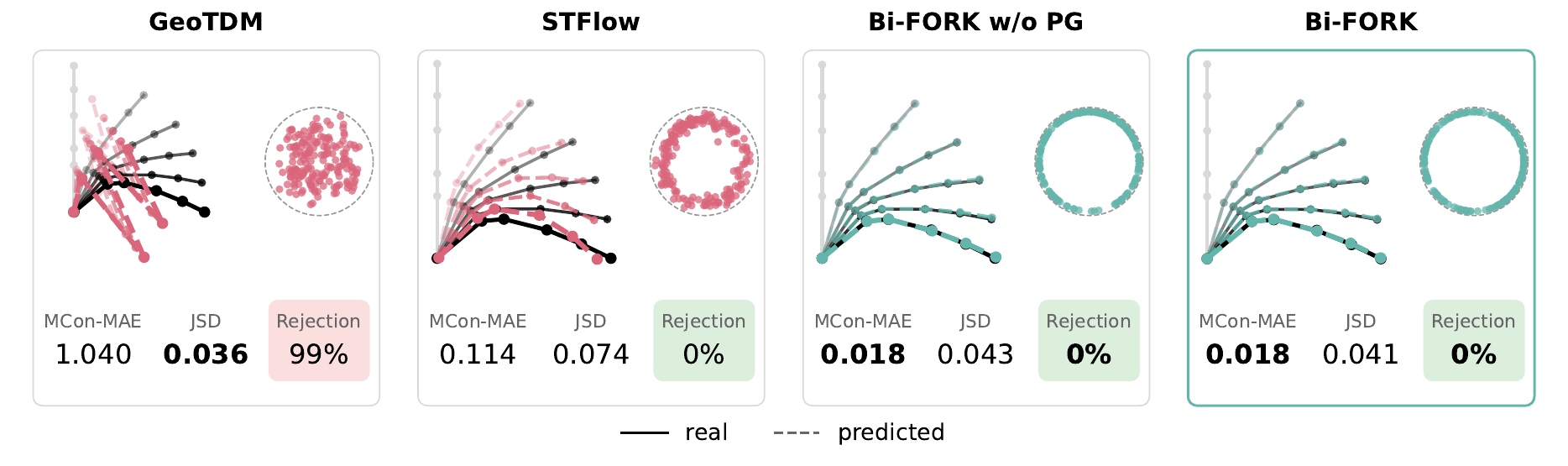}
    \caption{Beam3D result comparison.
    Each panel shows an evaluated method, with an illustrative sample (colored) plotted over the actual trajectory (black). Their endpoints are shown against the full population and the ideal uniform dashed circle.
    Metrics are aggregated over all Beam3D test conditions using five random seeds.
    }
    \label{fig:beam3d_results}
\end{figure}

\paragraph{Bifurcating System and Learning Task}
The 3D buckling beam dataset represents multiple slender beams with a geometry $\bfX^1$, which varies between data points, under axial compression. Each beam is fixed at one end with an incrementally increasing downward displacement $d$ at the other. The input condition is therefore $\mathbf{c}=(\bfX^1,\boldsymbol{\ell})$.
Below a critical threshold, the beam stays in axial compression.
Beyond the threshold, the rotational symmetry breaks: the axisymmetric state remains an equilibrium but is no longer stable, and the beam buckles in an azimuthal direction $\phi \in [0, 2\pi]$, yielding a continuous \(SO(2)\) orbit of solutions.
Given the condition $\mathbf{c}$, the model is required to predict a representative set of $K$ trajectories $\{\widehat{\bfX}^{(k)}\}_{k=1}^K$, sampled from this continuous orbit.



\paragraph{Modeling}
To model the bifurcating system, we use the architecture described in Section~\ref{subsec:modeling_bf}.
In addition to tip displacement, we input per-node rotational stiffness and element-level attributes, including stiffness and length.
At inference, we use particle guidance to distribute the samples more evenly over the admissible post-buckling manifold $\mathcal{M}_{\mathbf{c}}$. We additionally evaluate \ourmethod~without particle guidance; we denote this variant \ourmethod~w/o PG.
We sample $180$ times to estimate the distribution accuracy.
Details on the chosen architecture parameters, such as latent dimension or hidden size, are provided in Appendix \ref{app:experimental_details}.

\paragraph{Metrics} In addition to the metrics described in \cref{subsec:metrics}, we report the rejection rate ($\%$): the fraction of generated trials whose mean Euclidean displacement from their matched mode exceeds a threshold. MCon-MAE and JSD are computed over all generated trials, including those that would have been rejected by the threshold.

\paragraph{Results}
Bifurcating dynamical systems are naturally formulated as interacting particle systems, motivating models that operate on sets of arbitrary size without retraining.
We compare against two such baselines: GeoTDM \citep{han2024geometric} and STFlow \citep{brinke2025flow}.
\cref{fig:beam3d_results} reports the MCon-MAE, JSD and rejection rate on the held-out test dataset.
Among the models, GeoTDM achieves the lowest JSD due to its built-in equivariance, but rejects $99\%$ of its samples, with the worst MCon-MAE.
STFlow reduces the trajectory rejection rate while maintaining low error and JSD values. Notably, \ourmethod{} substantially improves MCon-MAE with a $0\%$ trajectory rejection rate and achieves a highly competitive JSD, actually closing the gap with the equivariant GeoTDM model.
Visual inspection of the end-point distributions (\cref{fig:beam3d_results}) confirms the quantitative findings: GeoTDM's samples cluster off the target circle manifold, and STFlow samples are scattered noisily around it.
Both \ourmethod{} w/o PG and \ourmethod{} trace the target manifold accurately. \cref{tab:beam3d_results} summarizes performance across five random seeds, reporting the mean and standard deviation.

We also compare inference runtime, shown in \cref{tab:wall_clock_time}. \ourmethod{} offers significant runtime advantages over graph-based baselines. It generates 180 samples for a single condition in 3.95 seconds, making it approximately 18× faster than STFlow and 1,966× faster than GeoTDM.

\subsection{Mechanical Metamaterials}
\label{subsec:mm_exp}
\begin{figure}[ht]
    \centering
    \includegraphics[width=\linewidth]{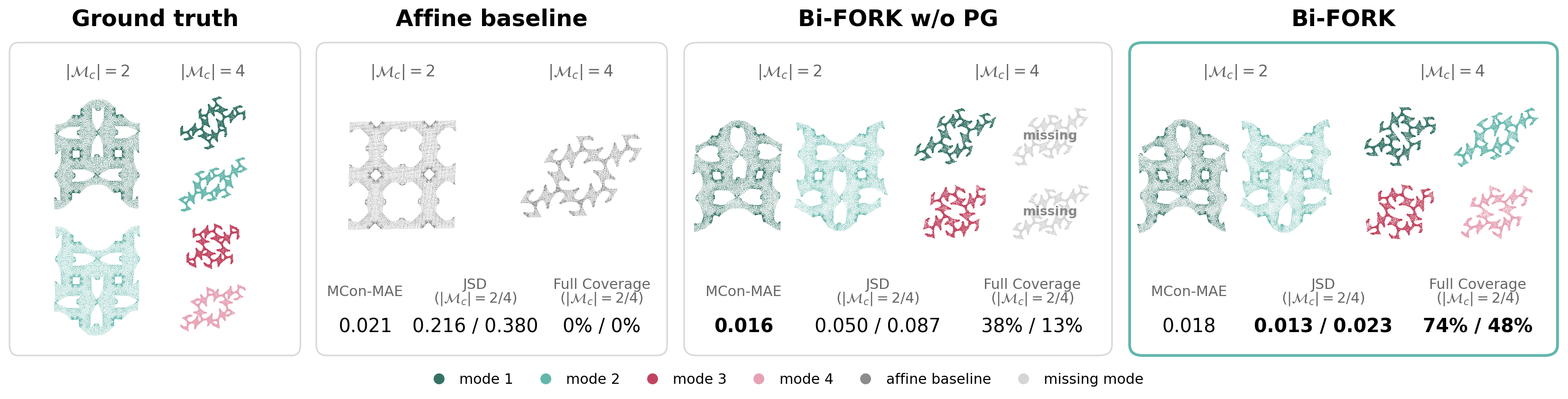}
    \caption{Mechanical metamaterials result comparison. Ground truth modes and generated samples are shown for representative conditions with $|\mathcal{M}_{\mathbf{c}}|=2$ and $|\mathcal{M}_{\mathbf{c}}|=4$ buckling modes. Colors indicate the mapped ground-truth mode; missing modes are shown in gray. Affine baseline results in mean prediction. Reported MCon-MAE is averaged over the full test set, while JSD and full coverage are reported for $|\mathcal{M}_{\mathbf{c}}|\in\{2,4\}$.
    Metrics reported over five random seeds on the test dataset.}
    \label{fig:mm_results}
\end{figure}
\paragraph{Bifurcating System and Learning Task}
The dataset of~\citet{hendriks2025wallpaper} contains buckling trajectories of periodic porous metamaterials spanning all 17 wallpaper groups. Depending on geometry and loading, symmetry breaking produces 1, 2, or 4 translation-related solutions, i.e., $|\mathcal{M}_{\mathbf{c}}|\in\{1,2,4\}$: one mode when no observable symmetry breaking occurs, and two or four modes after isotropy reduction, with four modes potentially arising after a secondary bifurcation.

The condition is therefore $\mathbf{c}=(\bfX^1,\boldsymbol{\ell})$ with $\boldsymbol{\ell}^t=\bfF^t$, with $\bfF^t$ the applied macroscopic deformation gradient at time $t$. Since $\bfF^t$ is known, the predicted $\{\widehat{\bfM}^t\}_{t=1}^T$ are the microfluctuation fields instead of the full deformation field.
The deformed coordinates are then reconstructed as $\widehat{\bfX}^t=\bfX^1(\bfF^t)^\top+\widehat{\bfM}^t$.
Unlike the beam, the mode distribution here is discrete: the model must predict the correct number of modes for a given condition, because it should generate each physically plausible branch in $\mathcal{M}_{\mathbf{c}}$.

\paragraph{Modeling}
Following the architecture in \cref{subsec:modeling_bf}, we use the first frame as the conditioning frame for each geometry.
Additionally, we add node type information and the crystallographic group.
Inference is executed similarly to the Beam3D dataset, with one difference for \ourmethod: when no bifurcation time point is detected ($|\mathcal{M}_{\mathbf{c}}| = 1$), the sampling algorithm discards the repulsive term in the diffusion process.
In this setting, we sample $16$ times to estimate the potential distribution for each condition.

\paragraph{Metrics} We employ the metrics from \cref{subsec:metrics}. Additionally, we report coverage ($\%$), defined as the percentage of conditions for which every ground-truth mode is represented by a valid generated sample.
We report JSD and coverage only for cases with $|\mathcal{M}_{\mathbf{c}}| \in \{2,4\}$. MCon-MAE and JSD are computed over all generated trials, including those that would have been rejected by the threshold.

\begin{wraptable}{r}{0.55\textwidth}
    \centering
    \vspace{-1.2em}
    \caption{Mode distributions of the two illustrated geometries in \cref{fig:mm_results} with $|\mathcal{M}_{\mathbf{c}}|\in\{2,4\}$. Each row aggregates five seeds, with $K=16$ samples per seed.}
    \label{tab:microstructures-sample-mode-distributions}
    \begin{adjustbox}{max width=\linewidth}
    \begin{tabular}{@{}llccccr@{}}
        \toprule
        \textbf{Sample} & \textbf{Method} & \textbf{$K$} & \textbf{Mode 1} & \textbf{Mode 2}
        & \textbf{Mode 3} & \textbf{Mode 4} \\
        \midrule
        \multirow{2}{*}{2 modes}
          & \ourmethod~w/o PG   & 16 & 80.0\% & 20.0\% & --     & --     \\
          & \ourmethod{} & 16 & 56.2\% & 43.8\% & --     & --     \\
        \midrule
        \multirow{2}{*}{4 modes}
          & \ourmethod{} w/o PG   & 16 & 18.2\% &  0.0\% & 72.7\% &  9.1\% \\
          & \ourmethod{} & 16 & 23.2\% & 15.4\% & 36.3\% & 25.1\% \\
        \bottomrule
    \end{tabular}
    \end{adjustbox}
    \vspace{-1.2em}
\end{wraptable}

\paragraph{Results}
GeoTDM and STFlow were not viable baselines for this problem setting because by default, both rely on a fully connected graph, which is infeasible with thousands of nodes. They can also use sparse connectivity based on a distance threshold, but then they would need a number of message-passing steps equal to the graph diameter to achieve all-to-all communication, which is infeasible too.
On Beam3D ($\leq 11$ nodes for $T=200$), both models showed no improvement and, being GNN-based, were already slow to train and sample from, requiring up to three orders of magnitude more inference time than Bi-FORK (\cref{tab:wall_clock_time}).
With roughly $160\times$ more nodes per timestep and all timesteps processed jointly (\cref{sec:introduction}), both models become computationally infeasible.
We therefore compare against the affine baseline, which propagates $\bfX^1$ using only the deformation gradient. \cref{fig:mm_results} shows that \ourmethod{} w/o PG already improves over the affine baseline in accuracy and in distributional coverage: the per-mode JSD shows that it recovers multiple modes rather than collapsing to a mean prediction.
With the full potential of our method, we further improve coverage, retrieving all four modes in roughly $50\%$ of the sampled test cases. \Cref{tab:results_mm} provides the quantitative summary, including standard deviation. \cref{tab:microstructures-sample-mode-distributions} shows the diversity improvement for the plotted samples with two and four modes, producing a nearly uniform distribution with \ourmethod.

Finally, we ablate the bifurcation head on Beam3D and mechanical metamaterials.
The results in Table~\ref{tab:bifurcation_classification} show that the learned bifurcation masks are predicted accurately on both datasets.
\subsection{Allen-Cahn}
\label{subsec:allencahn_exp}
\begin{figure}[ht]
    \centering
    \includegraphics[width=\linewidth]{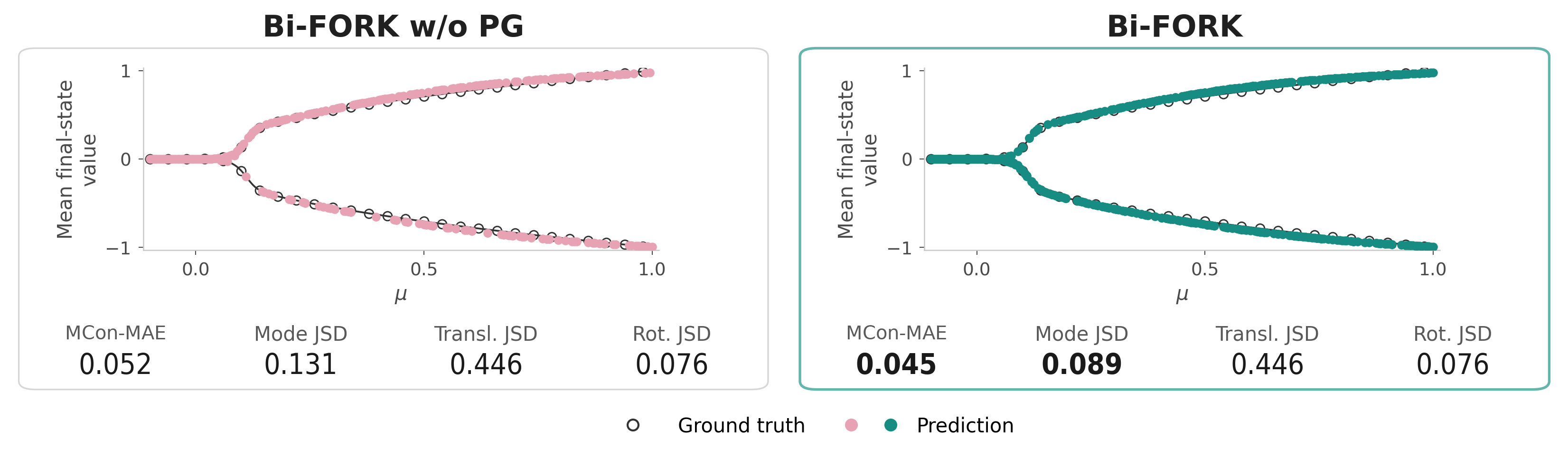}
    \caption{Allen-Cahn result comparison. Ground truth (black open circles) and predictions (teal and pink circles) are shown across a parameter sweep $\mu \in [-0.1, 1.0]$ at fixed $\epsilon = 0.1$. The vertical axis shows the final-frame mean value. We evaluate metrics on the test dataset and average over five random seeds; lower values indicate better performance.}
    \label{fig:ac_results}
\end{figure}

\paragraph{Bifurcating System and Learning Task}
The 3D Allen-Cahn dataset describes the evolution of a phase-field variable governed by the Allen-Cahn equation, which models phase separation in binary mixtures on a periodic domain.
Because all trajectories share the same initial homogeneous state, the condition is only $\mathbf{c} = (\epsilon, \mu)$, which is fixed over time. This initial state is invariant to $G_{\mathbf{c}}$, which contains phase exchange $u \mapsto -\,u$, periodic translations and the cubic point group.
The initial state can evolve into many patterns, which can be symmetry-related through $G_{\mathbf{c}}$ or morphologically distinct.
The number of morphologically distinct modes rises as $\epsilon^{-3}$: shrinking $\epsilon$ opens more unstable directions.

\paragraph{Modeling}
Given Allen-Cahn's grid structure, the first stage uses a ViT instead of the Perceiver-style encoder.
The periodic scalar field is compressed into a compact spatial-latent-token grid. Implementation details are in Appendix~\ref{app:experimental_details}.
The second stage follows \cref{subsec:modeling_bf}.
As input conditions, we only use the physical parameters $\epsilon$ and $\mu$; since the full trajectory is generated starting at $t=1$, no additional conditioning frame is needed.
The output is the full scalar-field rollout.
We apply symmetry-aware augmentations (periodic translations and sign flips) that match the domain's periodicity and point-group structure.
At inference, the sampling follows the previously described pipeline, with one exception: we apply particle guidance at every valid frame by setting $\widehat{b}^t=1$ throughout the sampling process. Therefore, we do not train a bifurcation classification head for this problem setting.

\paragraph{Metrics}
We extend the metrics described in \cref{subsec:metrics} to the Allen-Cahn solution space.
A raw distance between a generated field and any single reference is uninformative when the reference could equally be shifted, reflected, rotated, or sign-inverted and still represent the same solution.
We therefore first match each generated trial to the closest branch among a predefined set of ground-truth solutions using an FFT-based descriptor invariant to these symmetries, then align it to that branch via an exhaustive search over the point group.
Full implementation details are given in Appendix \ref{app:evaluation_allen_cahn}.

Estimating the JSD on the joint distribution over (branch, rotoreflection, shift) is infeasible, as the number of possible combinations exceeds the number of available samples. 
We therefore split the JSD metrics into three separate components:
\textbf{mode-coverage JSD} over the predefined solutions (including their sign-inverted counterparts);
a \textbf{translation JSD} that similarly fixes a branch and tests whether the recovered periodic shift is uniform;
and a \textbf{rotoreflection JSD} that fixes a single matched branch and tests whether the recovered rotations and reflections relative to it are uniform.
MCon-MAE compares the full predicted trajectory with its relaxed counterpart, obtained by a Gauss–Newton projection onto the Allen–Cahn PDE.
It thus measures the distance to a physically valid solution rather than to precomputed references.

\begin{wrapfigure}{r}{0.42\textwidth}
\centering
    \vspace{-1.8em}
    \includegraphics[width=0.9\linewidth]{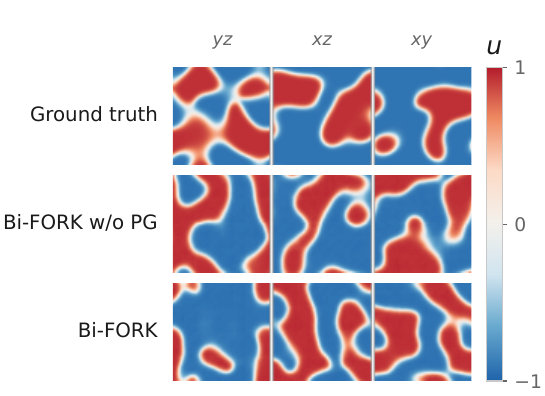}
    \caption{Visual comparison of ground truth vs predicted samples for $\mu = 0.86$ and $\epsilon = 0.094$.}
    \label{fig:example_results_ac}
\end{wrapfigure}

\paragraph{Results}
\Cref{fig:ac_results} summarizes the results, and \cref{app:table_ac_results} reports the mean and standard deviation over five seeds.
\ourmethod~reduces the MCon-MAE from 0.052 (without particle guidance) to 0.045, showing that the generated trajectories require smaller corrections to satisfy the PDE.
It also lowers the mode JSD from 0.131 to 0.089, reflecting improved coverage of the solution branches.
The translation and rotoreflection JSDs are identical for both variants.
To illustrate mode coverage, we fix $\epsilon = 0.1$ and vary only $\mu$.
Without particle guidance, the model samples mostly from the positive branch and misses several solutions on the negative branch.
\ourmethod~samples both branches with similar density over the full range of $\mu$. \Cref{fig:example_results_ac} shows samples from both methods next to the ground truth. Even when the samples differ, they showcase valid solutions.

\section{Discussion}
\label{sec:Discussion}

\paragraph{Limitations} Our experiments show that \ourmethod~applies to a diverse set of bifurcating systems; however, some limitations remain. First, particle guidance can trade accuracy for diversity: the repulsion acts on all samples, including those already in the same valid mode, and can push them off the solution manifold. Second, the Allen–Cahn evaluation is necessarily incomplete. The full distribution over morphologies is unavailable, so the mode-coverage JSD relies on four simulated realizations per condition and their sign flips. In addition, separate JSDs for modes, translations, and rotoreflections do not guarantee coverage of the joint distribution. Third, although symmetry is central to the problem, the model learns it from data and augmentation rather than enforcing it architecturally.

\paragraph{Conclusion}
We introduce \ourmethod, a framework for learning symmetry-breaking bifurcations. It maps solution multiplicity via symmetry orbits and bifurcation time. The task is framed as learning a conditional distribution over complete trajectories using latent flow matching and particle guidance. We evaluate models using symmetry-aware metrics that detect mode collapse. Across buckling beams, mechanical metamaterials, and Allen-Cahn phase separation, \ourmethod~recovers multimodal solutions at discretizations up to 260,000 points, several orders of magnitude beyond prior work. Because no existing stochastic method scales to this size, we evaluate the largest systems against a deterministic affine baseline and our own ablations. Ultimately, this opens generative modeling to large-scale bifurcating physical systems.

\subsection*{Acknowledgments}
We wish to thank Andreas Fürst for helpful discussions and feedback.
The ELLIS Unit Linz, the LIT AI Lab, the Institute for Machine Learning, are supported by the Federal State Upper Austria.
We thank the European Union’s Horizon Europe research and innovation programme under grant agreement number 101214398 (ELLIOT) and the Austrian Science Fund (FWF) 10.55776/COE12.
We thank Audi AG, Silicon Austria Labs (SAL), Merck Healthcare KGaA, GLS (Univ. Waterloo), T\"{U}V Holding GmbH, Software Competence Center Hagenberg GmbH, dSPACE GmbH, TRUMPF SE + Co. KG.
We thank the projects FWF AIRI FG 9-N (10.55776/FG9), AI4GreenHeatingGrids (FFG-899943), FWF Bilateral Artificial Intelligence (10.55776/COE12).

\subsection*{AI use statement}
In this work, we used generative AI tools for three tasks requiring disclosure: implementing methods, cleaning and reformatting datasets, and formulating mathematical claims, where AI served only as a source of inspiration. We used Codex to implement the baseline methods and supporting evaluation code in \cref{sec:experiments}; two authors reviewed, verified, and tested this code, and we implemented our proposed method ourselves. We used Codex and Claude to clean and reformat the datasets and checked the processed data against the original sources. Codex and Claude also offered ideas for the mathematical claims in \cref{sec:bifurcation_learning_problem}, but the final claims were formulated and verified by the authors. We have not used generative AI tools to generate synthetic datasets; to develop the problem setting, problem framing, or proposed architecture; to propose or refine hypotheses; to design or give feedback on experiments; or to interpret results. Translation and qualitative or thematic data analysis are not applicable to this work.

Additionally, we used generative AI tools to identify relevant literature, to polish the manuscript for readability, and to make cosmetic improvements to figures. We did not use them for other recommended-disclosure tasks such as drafting text, suggesting hyperparameters, identifying research gaps, formatting references, or proposing a title. We have reviewed all AI-assisted work: suggested papers were read by the authors before being cited, text edits were checked to ensure that the technical content and claims were unchanged, and figure edits were limited to styling, with the underlying numbers never modified by AI. We take responsibility for the final content of this work, including text, claims, and artifacts produced with the aid of generative AI.

\subsection*{Reproducibility statement}
We provide a detailed description of the datasets, including the evaluation splits used, in Appendix~\ref{app:datasets}. Appendix~\ref{app:experimental_details} also elaborates on the model architecture, training pipeline, loss functions, and hyperparameters for each dataset. Additionally we provide a detailed overview on the evaluation pipeline for each dataset in Appendix~\ref{app:evaluation_beam3D_mm} for Beam3D and mechanical metamaterials, and in Appendix~\ref{app:evaluation_allen_cahn} for the Allen-Cahn dataset. The code is available at: \url{https://github.com/ml-jku/Bi-FORK}.



{
\small
\bibliographystyle{plainnat}
\bibliography{refs}
}

\newpage
\appendix
\startcontents[appendix]
\section*{Appendix Contents}
\printcontents[appendix]{}{1}{\setcounter{tocdepth}{2}}
\clearpage

\section{Notation}
\label{app:notation}
The notation used throughout the paper is summarized in \cref{tab:notation}.

\begin{table}[h!]
    \centering
    \caption{Overview of used symbols and notations. Superscripts $t$ index trajectory frames, subscripts $\tau$ flow time, and parenthesized superscripts $(k)$ generated samples. Dataset-specific quantities are defined where they are used (\cref{sec:experiments,app:datasets}).}
    \label{tab:notation}
    \small
    \begin{tabular}{@{}l >{\raggedright\arraybackslash}p{0.575\linewidth} l@{}}
        \toprule
        Symbol & Meaning & Space \\
        \midrule
        \multicolumn{3}{@{}l}{\textit{Data and trajectories}}\\
        $N$, $n$                  & number of spatial locations (nodes); node index & $\dN$ \\
        $T$, $t$                  & number of timesteps; timestep index & $\dN$ \\
        $D$, $C$                  & spatial dimension; channels of the response & $\{2,3\}$, $\{1,D\}$ \\
        $\bfX^1$, $\bfX^t$        & reference geometry; deformed coordinates at timestep $t$ & $\dR^{N\times D}$ \\
        $\bfM^t$                  & response (signal field) at timestep $t$ & $\dR^{N\times C}$ \\
        $\bfM^{1:T}$              & response trajectory $(\bfM^1,\dots,\bfM^T)$ & $\dR^{T\times N\times C}$ \\
        \midrule
        \multicolumn{3}{@{}l}{\textit{Conditions, modes, and symmetry}}\\
        $\mathbf{c}$, $\mathcal{C}$ & condition $(\bfX^1,\boldsymbol{\ell})$, or $(\epsilon,\mu)$ for Allen-Cahn; set of conditions & \\
        $\boldsymbol{\ell}$       & control protocol (loading schedule) $(\boldsymbol{\ell}^1,\dots,\boldsymbol{\ell}^T)$ & \\
        $\mathcal{M}_{\mathbf{c}}$, $\mathcal{M}_{\mathbf{c}}^t$ & mode set (admissible trajectories); responses at timestep $t$ & $\subset\dR^{T\times N\times C}$ \\
        $G_\mathbf{c}$            & symmetry group of the condition & \\
        $\bfM_\mathbf{c}^t$       & representative response at timestep $t$ & $\in\mathcal{M}_{\mathbf{c}}^t$ \\
        $H_\mathbf{c}^t$          & isotropy subgroup of $\bfM_\mathbf{c}^t$ & $\le G_\mathbf{c}$ \\
        $t_b$                     & bifurcation time: first strict decrease of $H_\mathbf{c}^t$ ($\infty$ if none) & $\{2,\dots,T\}\cup\{\infty\}$ \\
        \midrule
        \multicolumn{3}{@{}l}{\textit{Distributions}}\\
        $p(\bfM^{1:T}\mid\mathbf{c})$ & ground-truth distribution of trajectories, supported on $\mathcal{M}_{\mathbf{c}}$ & \\
        $p_{\theta_{\mathrm{flow}}}(\bfM^{1:T}\mid\mathbf{c})$ & learned distribution (latent flow model, decoded) & \\
        \midrule
        \multicolumn{3}{@{}l}{\textit{Latent representation}}\\
        $\mathcal{E}_{\phi_{\mathrm{enc}}}$, $\mathcal{D}_{\phi_{\mathrm{dec}}}$ & encoder $(\bfX^1,\bfM^t)\mapsto\bfZ^t$; decoder $(\bfZ^t,\bfX^1)\mapsto\widehat{\bfM}^t$ & \\
        $\bfZ^t$                  & latent state at timestep $t$ & $\dR^{L\times D_z}$ \\
        $L$, $D_z$                & number of latent tokens; token dimension & $\dN$ \\
        \midrule
        \multicolumn{3}{@{}l}{\textit{Latent flow matching}}\\
        $\tau$                    & flow time ($0$: noise, $1$: data) & $[0,1]$ \\
        $\bfZ_1$                  & clean latent trajectory $\bfZ^{1:T}$ & $\dR^{T\times L\times D_z}$ \\
        $\boldsymbol{\xi}$        & Gaussian noise & $\mathcal{N}(0,I)$ \\
        $\bfZ_\tau$               & noisy latent trajectory $\tau\bfZ_1+(1-\tau)\boldsymbol{\xi}$ & $\dR^{T\times L\times D_z}$ \\
        $T_{\mathrm{obs}}$        & number of observed (conditioning) frames & $\{0,\dots,T\}$ \\
        $f_{\theta_{\mathrm{flow}}}$, $\theta_{\mathrm{flow}}$ & predictor $(\bfZ_\tau,\tau,\bfZ^{1:T_{\mathrm{obs}}},\mathbf{c})\mapsto\widehat{\bfZ}_1$; its parameters & \\
        $\widehat{\bfZ}_1$        & predicted clean latent trajectory & $\dR^{T\times L\times D_z}$ \\
        $\widehat{v}_{\theta_{\mathrm{flow}}}$ & velocity field of the probability-flow ODE, induced by $\widehat{\bfZ}_1$ & $\dR^{T\times L\times D_z}$ \\
        \midrule
        \multicolumn{3}{@{}l}{\textit{Sampling and particle guidance}}\\
        $K$, $k$                  & number of jointly generated samples; sample index & $\dN$ \\
        $\bfZ_\tau^{(k)}$         & latent state of sample $k$ & $\dR^{T\times L\times D_z}$ \\
        $p^t$, $\widehat{\bfb}$   & predicted post-bifurcation probability; mask $\widehat{b}^t=\mathbbm{1}[p^t\ge 1/2]$ & $[0,1]$, $\{0,1\}^T$ \\
        $\mathrm{force}^{(k)}$    & repulsive force on sample $k$ & $\dR^{T\times L\times D_z}$ \\
        $w_{\mathrm{vel}}$, $w_{\mathrm{rep}}$ & weights of model velocity and repulsion & $\dR_{\ge 0}$ \\
        $\widehat{v}^{(k)}$       & guided velocity of sample $k$ & $\dR^{T\times L\times D_z}$ \\
        \midrule
        \multicolumn{3}{@{}l}{\textit{Training and evaluation}}\\
        $\mathcal{L}_{\mathrm{rec}}$, $\mathcal{L}_{\mathrm{flow}}$, $\mathcal{L}_{\mathrm{bif}}$ & reconstruction, latent-flow, and bifurcation losses & \\
        $b^t$, $r^t$              & bifurcation label $\mathbbm{1}[t\ge t_b]$; valid-frame indicator & $\{0,1\}$ \\
        $\widehat{\bfX}_{\mathbf{c}}^{(k)}$, $\bfX_{\mathbf{c},m}$ & generated trajectory $k$; ground-truth trajectory of mode $m$ & \\
        $m_\star^{(k)}$           & ground-truth mode matched to sample $k$ & \\
        MCon-MAE, JSD             & mode-conditioned MAE; Jensen--Shannon divergence & $\dR_{\ge 0}$ \\
        \bottomrule
    \end{tabular}
\end{table}

\newpage

\section{Additional bifurcation context}
\label{app:bifurctime_context}
\paragraph{Estimating the bifurcation time}
In the datasets, $t_b$ is obtained from simulated trajectories rather than from the isotropy subgroups directly.
For two trajectories $i,j$ under the same condition, we compute the pairwise divergence $\mathrm{d}_{ij}^t=\lVert \bfM_i^t-\bfM_j^t\rVert$ (the load-prescribed part cancels) and take the first timestep at which it exceeds a small threshold $\vartheta>0$.
Before the bifurcation, the equilibrium is unique and $G_\mathbf{c}$-invariant, so all trajectories coincide; afterwards, trajectories on distinct branches separate.
For the quasi-static datasets, this estimate therefore recovers $t_b$ up to the resolution $\vartheta$.
For Allen-Cahn, where small noise perturbs the shared homogeneous initial state, it marks the onset of macroscopic domain formation.

\paragraph{Non-bifurcating conditions}
If no timestep satisfies $H_\mathbf{c}^t\subsetneq H_\mathbf{c}^{t-1}$, the condition admits a single mode, 
and we set $t_b:=\infty$.
The bifurcation labels are then $b^t=\mathbbm{1}[t\ge t_b]\equiv 0$, and the estimator above detects no separation.
Accordingly, the bifurcation head is trained to predict $\widehat{\bfb}=\mathbf{0}$ for such conditions, which deactivates the repulsion term in Eq.~\ref{eq:guided-velocity}. 

\paragraph{Secondary bifurcations}
A loading path can exhibit several isotropy drops $t_b^{(1)}<t_b^{(2)}<\cdots$, i.e., hierarchical bifurcations; the four-mode metamaterials arise from two such drops (\cref{fig:mode_4_figure}). Throughout, $t_b$ denotes the first one.

\paragraph{Choice of representative}
The definition of $t_b$ in \cref{subsec:bifurcating} refers to a representative response $\bfM_\mathbf{c}^t$ at every timestep.
Two responses on the same orbit have conjugate isotropy subgroups, $H_{g\cdot\bfM}=g\,H_{\bfM}\,g^{-1}$ for $g\in G_\mathbf{c}$, and conjugate subgroups are in general not nested.
The inclusion $H_\mathbf{c}^t\subsetneq H_\mathbf{c}^{t-1}$ is therefore to be read along a single trajectory: $\bfM_\mathbf{c}^t$ denotes frame $t$ of one fixed admissible trajectory $\bfM_\mathbf{c}^{1:T}\in\mathcal{M}_{\mathbf{c}}$.
With this convention, $t_b$ does not depend on the trajectory chosen.
Replacing $\bfM_\mathbf{c}^{1:T}$ by $g\cdot\bfM_\mathbf{c}^{1:T}=(g\cdot\bfM_\mathbf{c}^1,\dots,g\cdot\bfM_\mathbf{c}^T)$ conjugates every $H_\mathbf{c}^t$ by the same $g$; conjugation by a fixed element preserves inclusions and their strictness, so the first timestep at which $H_\mathbf{c}^t\subsetneq H_\mathbf{c}^{t-1}$ holds is unchanged.
For Allen-Cahn, where $\mathcal{M}_{\mathbf{c}}$ is a union of orbits, this defines $t_b$ per orbit; the estimator above applies unchanged.

\paragraph{Bifurcation, not chaos}
Unlike chaotic systems, where nearby trajectories diverge exponentially and long-run outcomes are unpredictable~\citep{ruelle1971nature, guckenheimer1983}, multistable systems branch into a \emph{small, structured} set of symmetry-related outcomes.
This makes the problem well posed for learning: the target is a multimodal distribution over a few valid branches, not an unpredictable attractor.

\paragraph{Relation to bifurcation theory}
For the mechanical datasets, $t$ counts load increments, so the ``bifurcation time'' is really a bifurcation \emph{load}: the classical setting of equilibrium bifurcation theory, where a solution branch splits once the equilibrium stops being unique---mechanically, where the tangent stiffness matrix becomes singular~\citep{koiter1945, thompson1973, antman2005, crandall1971, kielhofer2012, seydel2010}.
In symmetric problems, the new equilibria have strictly less symmetry and appear in symmetry-related copies, which is exactly the isotropy reduction defining $t_b$ in \cref{subsec:bifurcating} and the content of equivariant bifurcation theory~\citep{golubitsky1988, golubitsky2002}.
Allen-Cahn is instead genuinely time-dependent and covered by the dynamical version of the theory~\citep{guckenheimer1983, kuznetsov2004}.
We consider the generic supercritical scenario in which branches grow continuously out of the symmetric state; sudden jumps (subcritical or imperfect buckling) are out of scope.

\section{Datasets}
\label{app:datasets}

\subsection{Beam3D}
The dataset describes a 3D beam-buckling process, containing simulations with $3$ to $11$ nodes. In the problem setting, the beam is fixed at the base and aligned along the $z$-axis in its undeformed state, with a specified downward displacement applied incrementally to the tip. This displacement sequence is the control protocol $\boldsymbol{\ell}=(\boldsymbol{\ell}^1,\dots,\boldsymbol{\ell}^T)$ in the condition $\mathbf{c}=(\bfX^1,\boldsymbol{\ell})$.
Below a critical displacement threshold, the beam undergoes pure axial compression; beyond it, the system loses stability and deflects in any direction $\phi$ drawn uniformly from $[0, 2\pi]$ in the $x$-$y$-plane.
As every deflection direction is equally probable, the system admits $|\mathcal{M}_{\mathbf{c}}| = \infty$ distinct post-buckling modes under identical loading.

\paragraph{Dataset generation} The beam contains $N_{\mathrm{seg}}$ compressible elements, each characterized by a reference length $L_i^{\mathrm{beam}}$, axial stiffness $k_i^{\mathrm{ax}}$, and rotational spring constant $k_i^{\mathrm{rot}}$ resisting bending between neighboring elements.
The mechanical state is parametrized by axial strains $e_i = \ln(L_i^{\mathrm{def}} / L_i^{\mathrm{beam}})$, where $L_i^{\mathrm{def}}$ is the deformed element length, and $q_i$ represents the bending angle from the vertical. The resulting node positions are expressed relative to the reference geometry as the displacement trajectory $\{\bfM^t\}_{t=1}^T$ used by the model.
Equilibrium configurations minimize the strain energy
\begin{equation}
 E_{\mathrm{beam}}(\mathbf{q},\mathbf{e}) = \frac{1}{2}\left[k_1^{\mathrm{rot}}q_1^2 + \sum_{i=2}^{N_{\mathrm{seg}}} k_i^{\mathrm{rot}}(q_i-q_{i-1})^2\right] + \frac{1}{2}\sum_{i=1}^{N_{\mathrm{seg}}} L_i^{\mathrm{beam}} k_i^{\mathrm{ax}} e_i^2
\end{equation}
subject to $d = \sum_{i=1}^{N_{\mathrm{seg}}}L_i^{\mathrm{beam}}[1-\exp(e_i)\cos q_i]$, enforced via a Lagrange multiplier.
The dataset contains $997$ trajectories of up to $T=200$ timesteps, simulated to full deflection $d$, with system parameters $N_{\mathrm{seg}}$, $L_i^{\mathrm{beam}}$, $k_i^{\mathrm{rot}}$, $k_i^{\mathrm{ax}}$, and $\phi$ sampled across instances. The sampling distributions are summarized in \cref{tab:beam_sampling_summary}.

\begin{table}[ht]
    \centering
    \caption{Sampling distributions across used parameters.}
    \label{tab:beam_sampling_summary}
    \small
    \begin{tabular}{lc p{4cm}}
        \toprule
        \textbf{Parameter} & \textbf{Range} & \textbf{Add. Information} \\
        \midrule
        $N_{\mathrm{seg}}$          & 3–11 nodes (2–10 segments)   & per trajectory \\
        $L_i^{\mathrm{beam}}$       & log-uniform[$0.5, 2.0$]      & per segment \\
        $k_i^{\mathrm{ax}}$         & log-uniform[$0.5, 2.0$]      & per segment \\
        $k_i^{\mathrm{rot}}$        & log-uniform[$0.5, 2.0$]      & per node \\
        $\phi$                      & uniform on [$0, 2\pi$)        & per trajectory \\

        \bottomrule
    \end{tabular}
\end{table}

The dataset is well-suited to our analysis since the buckling direction $\phi$ is continuous, yielding infinitely many symmetry-related modes under the same loading condition.

\begin{figure}[ht]
    \centering
    \includegraphics[width=0.8\linewidth]{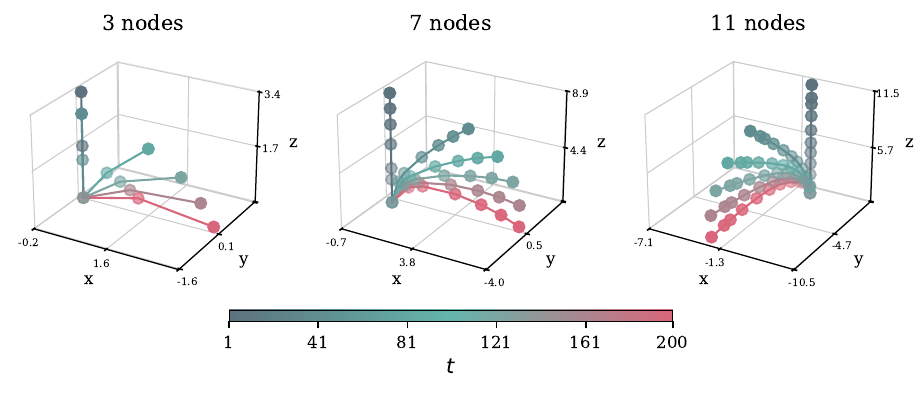}
    \caption{Example trajectories from the 3D beam dataset with different buckling directions $\phi$ for diverse node counts over timestep $t$.}
    \label{fig:beam3d_modes}
\end{figure}

\paragraph{Examples}
\Cref{fig:beam3d_modes} visualizes representative 3D beam trajectories for increasing node counts, showing how the initially vertical beam progressively buckles into different directions over time.

\paragraph{Dataset Split}
\Cref{tab:dataset_split_beam3d} illustrates the distribution across different data splits.
We consider a single sample per condition in the table, whereas during training the full range of buckling directions, $ \phi \in [0, 2\pi]$, is applied across the dataset.
\begin{table}[ht]
    \centering
    \caption{Summary of 3D beam dataset splits.}
    \label{tab:dataset_split_beam3d}
    \begin{tabular}{lcc}
        \toprule
        \textbf{Split} & \textbf{Trajectories} &  \textbf{Share (\%)} \\
        \midrule
        Training    & 697 & 70 \\
        Validation  & 150 & 15 \\
        Test        & 150 & 15 \\
        \midrule
        \textbf{Total} & 997 & 100 \\
        \bottomrule
    \end{tabular}
\end{table}

\subsection{Mechanical Metamaterials}
\label{app:mm_dataset}
For the second problem setting, we use the publicly available dataset of ~\citet{hendriks2025wallpaper}\footnote{\url{https://zenodo.org/records/18172947}}. It represents buckling trajectories of 2D porous elastic metamaterials, covering all 17 wallpaper groups. Each geometry is simulated on a $2\times2$ representative volume element under 12 loading paths, giving 12,240 trajectories from 1,020 geometries, with mesh sizes between 1,000 - 36,000 per timestep.
In our notation, the prescribed loading path is the control protocol $\boldsymbol{\ell}$; for this dataset each increment is a known deformation gradient, $\boldsymbol{\ell}^t=\bfF^t$. The trajectories contain up to $T = 32$ timesteps, but can terminate earlier if the system runs into contact and reaches a steady state.


For the underlying dataset the deformed configuration at timestep $t$ is obtained by combining the macroscopic deformation with the displacement field $\bfM^t$:
\begin{align}
    \bfX^t = \bfX^1 (\bfF^t)^\top + \bfM^t.
\end{align}
The deformation gradient is constructed by linearly interpolating from the identity matrix $\mathbf{I}$ to a prescribed terminal deformation gradient $\bfF_{T}$:
\begin{equation}
    \bfF^t = \mathbf{I} + \frac{t-1}{T-1}(\bfF_{T} - \mathbf{I}), \quad t \in \{1, \dots, T\}.
\end{equation}

The dataset is well-suited to our analysis because geometries can end in $|\mathcal{M}_{\mathbf{c}}| \in \{1,2,4\}$ distinct post-buckling modes under the same condition. Before bifurcation, symmetry-related modes should remain close in latent space; after bifurcation, the model must represent distinct modes without losing their shared trajectory-level structure.

\begin{figure}[h!]
    \centering
    \begin{subfigure}{0.48\linewidth}
        \centering
        \includegraphics[width=\linewidth]{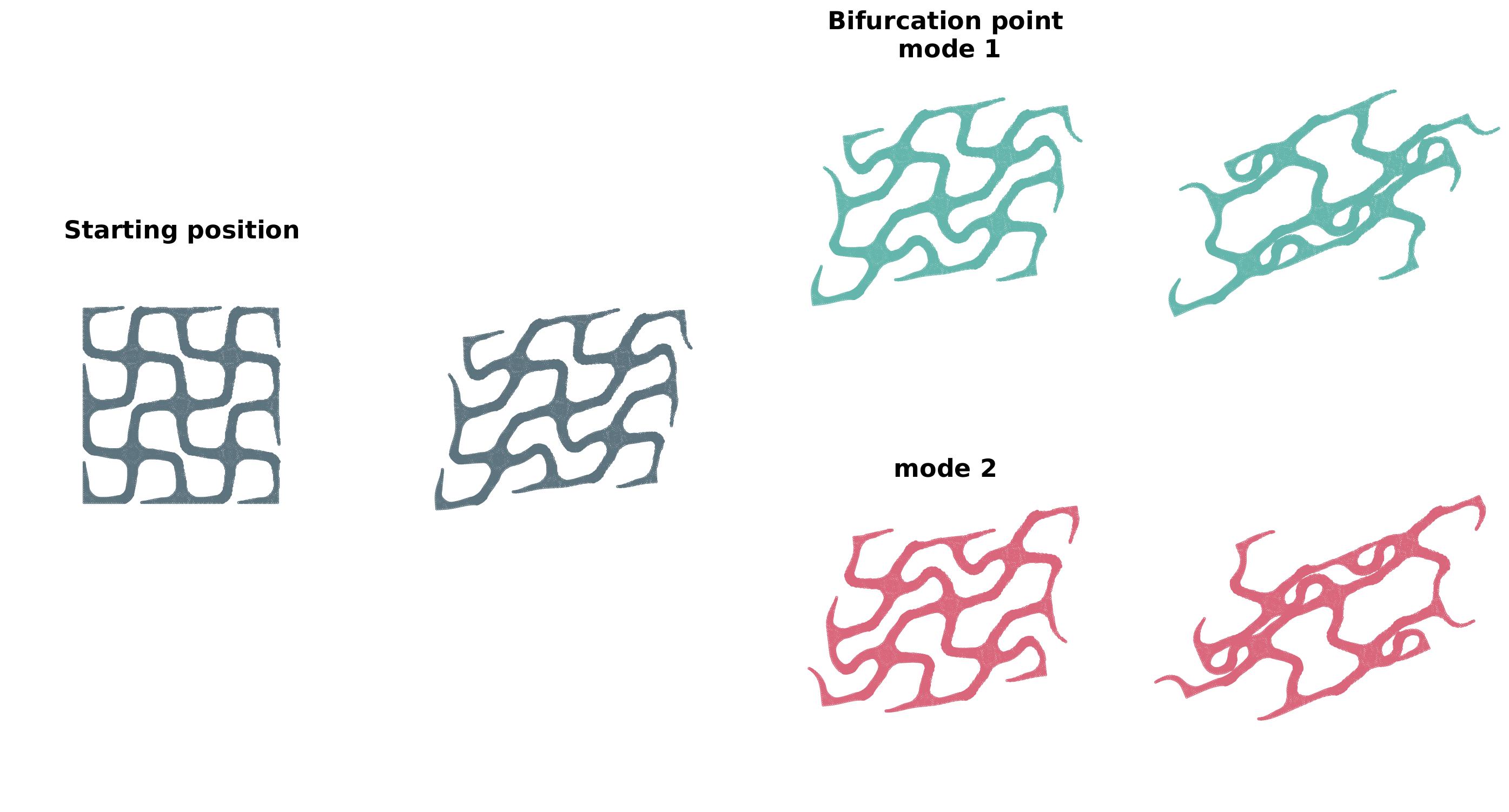}
        \caption{Example trajectory with one bifurcation point resulting in two modes.}
        \label{fig:mode_2_figure}
    \end{subfigure}
    \hfill
    \begin{subfigure}{0.48\linewidth}
        \centering
        \includegraphics[width=\linewidth]{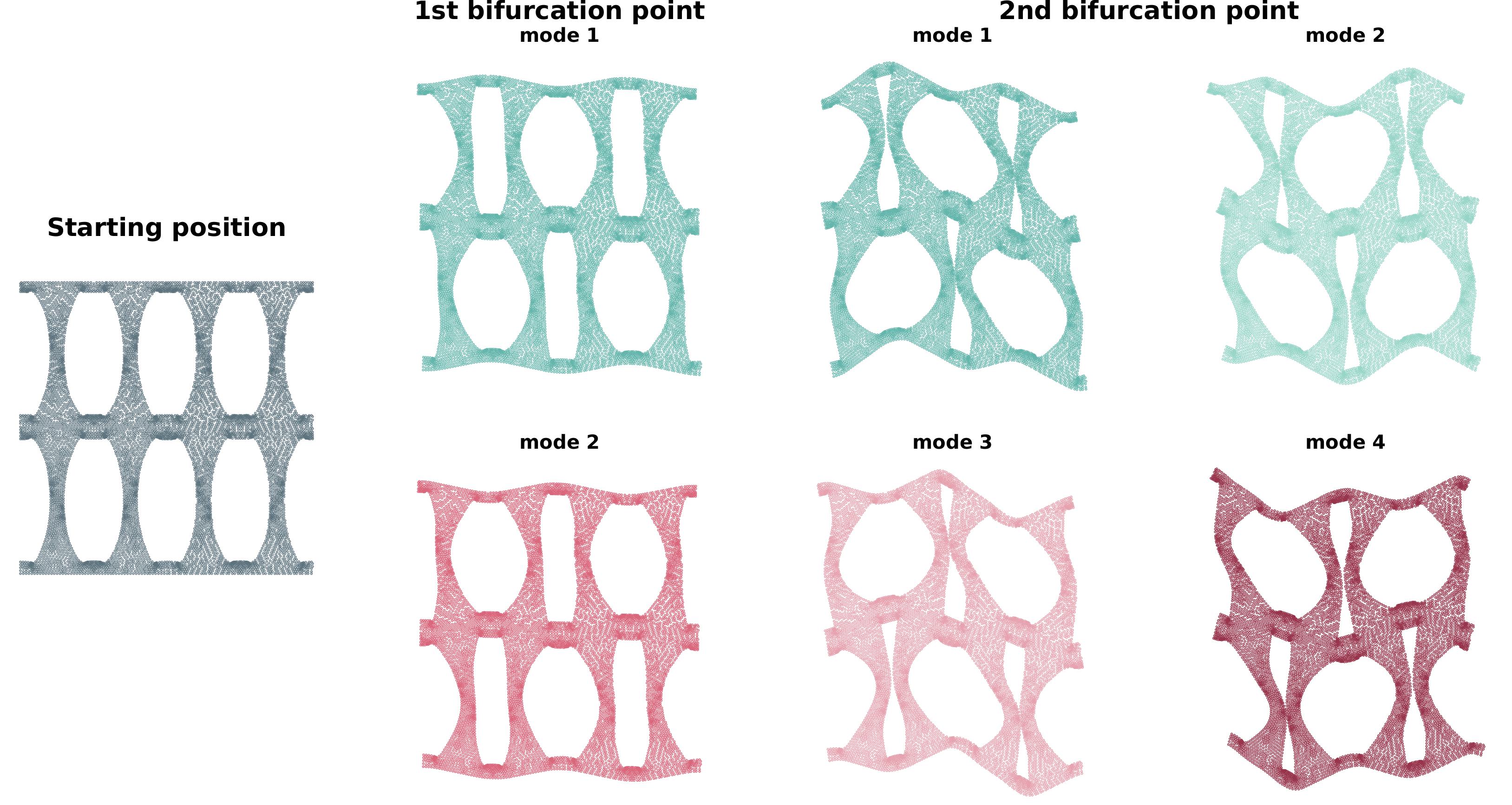}
        \caption{Example trajectory with two bifurcation points resulting in four modes.}
        \label{fig:mode_4_figure}
    \end{subfigure}
    \caption{Example mechanical metamaterials trajectories with distinct post-buckling mode structures.}
    \label{fig:mm_mode_examples}
\end{figure}

\paragraph{Examples}
\Cref{fig:mm_mode_examples} summarizes two examples of the metamaterials dataset with distinct modes.
\Cref{fig:mode_2_figure} shows a trajectory with 2 modes.
The starting configuration and the last timestep before buckling are displayed on the left in gray. Under progressive loading, the structure branches into two distinct post-buckling configurations: modes 1 and 2, which, in this case, are mirror reflections of one another. Both modes are shown at two successive timesteps, capturing how the transformed pattern becomes more pronounced with increasing load.

\Cref{fig:mode_4_figure} shows a trajectory with \textit{4 modes}. Starting from the undeformed configuration on the left, the structure first reaches a bifurcation point where two modes emerge. By moving toward the terminal deformation gradient $\bfF_{T}$, a second bifurcation point is reached, resulting in four distinct modes. This hierarchical branching behavior demonstrates how repeated symmetry-breaking events can generate increasingly varied deformed configurations from a single initial geometry.

\textbf{Dataset Split}
To prevent leakage between subsets, we split the data by geometry: each of the 1,020 geometries is assigned to exactly one of the training, validation, or test sets.
All 12 loading paths of a geometry are kept together in the same subset, so no test geometry is seen during training.
We provide a detailed overview of the training, validation, and test split in \cref{tab:dataset_split_mechmat} and their mode distributions.
Overall, the validation and test data share a similar number of modes across categories.

\begin{table}[ht]
    \centering
    \caption{Summary of dataset splits.}
    \label{tab:dataset_split_mechmat}
    \begin{tabular}{lccccc}
        \toprule
        \textbf{Split} & \textbf{1 mode} & \textbf{2 modes} & \textbf{4 modes} & \textbf{Trajectories} &  \textbf{Share (\%)} \\
        \midrule
        Training    & 6{,}529 & 1{,}777 & 262 & 8{,}568 & 70 \\
        Validation  & 1{,}408 & 377 & 51 & 1{,}836 & 15 \\
        Test        & 1{,}391 & 354 & 91 & 1{,}836 & 15 \\
        \midrule
        \textbf{Total} & 9{,}328 & 2{,}508 & 404 & 12{,}240 & 100 \\
        \bottomrule
    \end{tabular}
\end{table}

\subsection{Allen-Cahn Dataset}
\label{app:allen_cahn_dataset}

\begin{wrapfigure}{r}{0.42\textwidth}
\centering
    \includegraphics[width=0.9\linewidth]{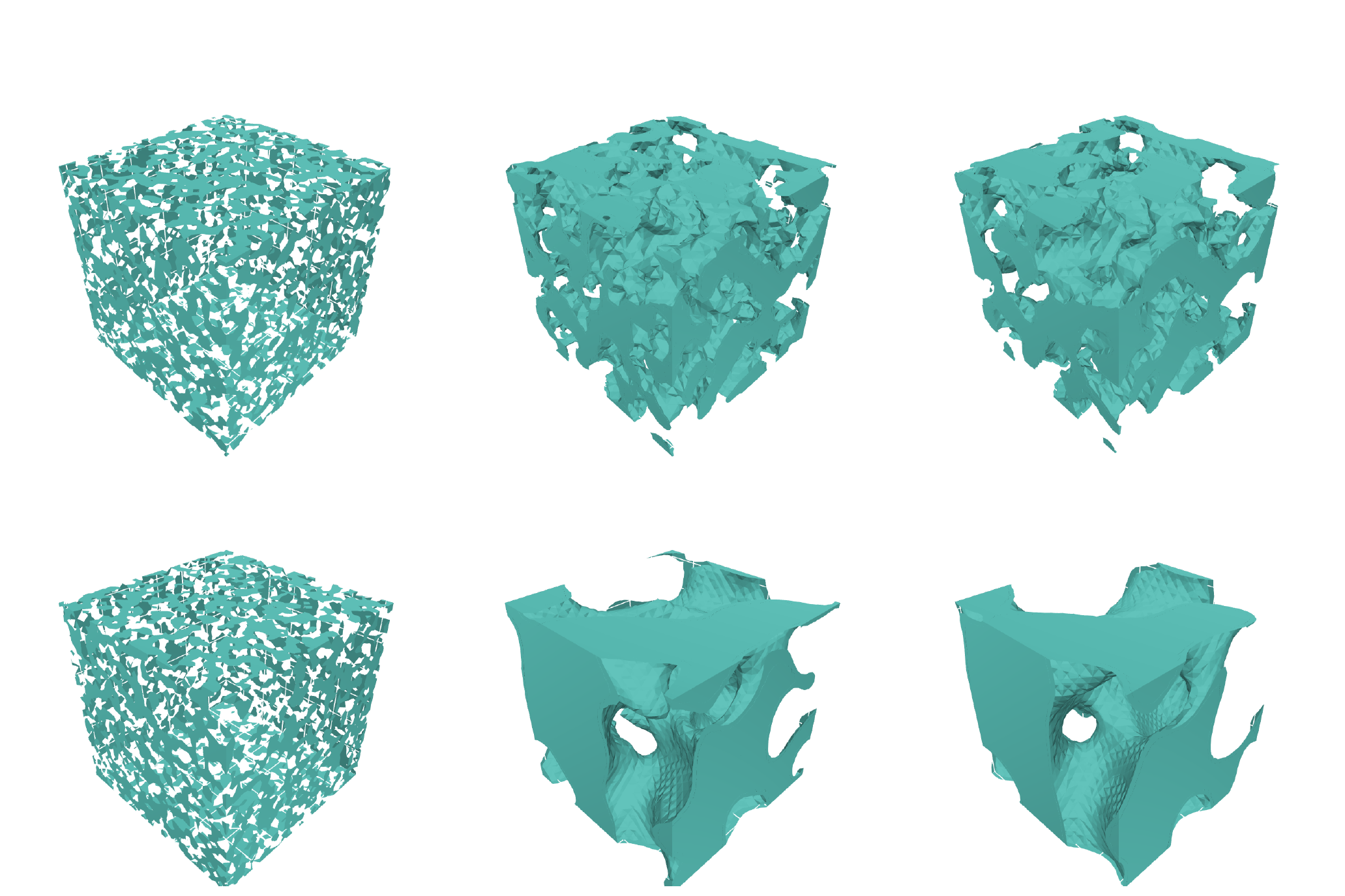}
    \caption{Allen-Cahn example for two distinct $\epsilon, \mu$ settings.}
    \vspace{-1.2em}
    \label{fig:overview_allen_cahn}
\end{wrapfigure}

The Allen-Cahn dataset describes phase separation in a three-dimensional periodic domain $\Omega$, see \cref{fig:overview_allen_cahn}.
The state is a scalar phase field $u(\mathbf{x}, t)\colon \Omega \times [0, 50] \to \mathbb{R}$ governed by
\begin{equation}
    \frac{\partial u}{\partial t}
    = \epsilon^2 \nabla^2 u - \left(u^3 - \mu u\right).
\end{equation}
Here, $\epsilon$ sets the characteristic width and smoothness of the interfaces, and $\mu$ determines the local phase structure.
For $\mu>0$, small perturbations of the mixed state grow and form spatial domains approaching $u=\pm\sqrt{\mu}$.
A small subset of the data uses $\mu < 0$, a non-separating regime in which the homogeneous state is stable.
We discretize $u$ on an $N_x \times N_y \times N_z$ grid and denote the discrete state at timestep $t$ by $\mathbf{u}^t \in \mathbb{R}^{N_x \times N_y \times N_z}$.

\paragraph{Dataset generation}
We generated $1{,}000$ parameter conditions on a $64^3$ grid over the periodic domain $[0,2\pi]^3$.
For every condition, $\epsilon$ was sampled log-uniformly as $\epsilon=10^z$ with $z\sim\mathcal{U}(-1.5,-1.0)$, i.e. $\epsilon\in[10^{-1.5},10^{-1}]\approx[0.0316,0.1]$.
Of the sampled $\mu$ values, $98\%$ were drawn uniformly from $[0.05,1.0]$ and $2\%$ from $[-0.1,0]$.

All trajectories start from the same homogeneous state $u^0(\mathbf{x}) \equiv 0$.
However, because this state is a valid solution but unstable, uniform random noise with an amplitude of 0.1 is added to the initial state in order to perturb it away from the unstable all-zero solution.
The equation is integrated for $5{,}000$ steps with $\Delta t=0.01$, using a semi-implicit Fourier-spectral scheme: the nonlinear reaction term is evaluated explicitly, whereas the diffusion term is solved implicitly by pointwise division in Fourier space.
Periodic boundary conditions are applied along all three axes.
The initial state and every $100$\textsuperscript{th} integration step are retained, giving $T=51$ snapshots per trajectory over the time interval $[0,50]$.
Consequently, the stored phase-field tensor before splitting has shape $4{,}000\times51\times64\times64\times64$.

\paragraph{Modes}
For Allen-Cahn the control protocol is constant, $\boldsymbol{\ell}^t\equiv(\epsilon,\mu)$.
For each sampled $(\epsilon,\mu)$ condition, four trajectories were simulated from the shared homogeneous initial state.
These trajectories share the same physical parameters and provide samples of different phase-separation morphologies under the same condition $\mathbf{c}$.
In addition, the Allen-Cahn equation is invariant under the analytic transformation $u\mapsto-u$.
Applying this sign flip to each realization provides its phase-exchanged counterpart, yielding eight candidate modes per condition (four simulated realizations and their four sign-reversed counterparts).

\paragraph{Dataset Split}
The data are split at the level of parameter conditions rather than individual trajectories.
Thus, all four trajectories belonging to the same $(\epsilon,\mu)$ pair remain in the same subset, preventing information leakage between training and evaluation.
The conditions were shuffled and divided into $70\%$ training, $15\%$ validation, and $15\%$ test data, as summarized in Table~\ref{tab:dataset_split_allen_cahn}.

\begin{table}[ht]
    \centering
    \caption{Summary of the Allen-Cahn dataset splits. Each condition contains
    four simulated trajectories and up to eight modes after analytic sign
    augmentation.}
    \label{tab:dataset_split_allen_cahn}
    \begin{tabular}{lcccc}
        \toprule
        \textbf{Split} & \textbf{Conditions} & \textbf{Simulated trajectories}
        & \textbf{Candidate modes} & \textbf{Share (\%)} \\
        \midrule
        Training   & 700   & 2{,}800 & 5{,}600 & 70 \\
        Validation & 150   & 600     & 1{,}200 & 15 \\
        Test       & 150   & 600     & 1{,}200 & 15 \\
        \midrule
        \textbf{Total} & 1{,}000 & 4{,}000 & 8{,}000 & 100 \\
        \bottomrule
    \end{tabular}
\end{table}

\section{Experimental Details}
\label{app:experimental_details}

\subsection{Architecture}
\label{app:architecture}

\paragraph{Autoencoder}
The encoder $\mathcal{E}_{\phi_{\mathrm{enc}}}$ and decoder $\mathcal{D}_{\phi_{\mathrm{dec}}}$ form a Perceiver-style cross-attention autoencoder~\citep{jaegle2021perceiver, alkin2024universal, sestak2026lamslide} for the 2D mechanical metamaterials and Beam3D datasets.
The encoder maps $N$ node points into a fixed, learned latent array of $L$ tokens.
The decoder reconstructs per-point deformations via cross-attention from positional queries.
All nonlinear activations are GELU~\citep{hendrycks2016gaussian}, and we additionally embed input and query positions via the sine-cosine position embedding from transformers \citep{vaswani2017attention}.
A schematic of the architecture is provided in \cref{fig:autoencoder_architecture}; dataset-specific autoencoder hyperparameters are listed in \cref{tab:ae_hparams}.

The Allen-Cahn Autoencoder uses a 3D Vision Transformer (ViT3D).
The encoder patchifies the input voxel grid to $8\times8\times8$ patches, which are run through self-attention/MLP transformer blocks to get a $4\times4\times4$ latent representation.
The decoder mirrors the encoder where the latent representation is upsampled to the original voxel grid.

We ensure latent stability by normalizing the physical fields and conditioning parameters using dataset-specific statistics, while layer normalization is used throughout the attention-based encoders and latent flow network.
This stabilizes the latent representations without introducing an additional variational regularization term.

\paragraph{Flow Matching Model}
We employ a two-stage latent flow matching model.
In the first stage, an encoder maps each physical state to a compact latent representation
\begin{equation}
\bfZ^t=\mathcal{E}_{\phi_{\mathrm{enc}}}(\bfX^1,\bfM^t),
\end{equation}
while a decoder reconstructs the original state.

The latent representation of a complete trajectory is

\begin{equation}
\bfZ_1 :=
\left(
\bfZ^1,\ldots,\bfZ^T
\right)
\in \dR^{T\times L\times D_z}.
\end{equation}

In the second stage, the autoencoder is frozen and a Scalable Interpolant Transformer (SiT) \citep{ma2024sit, albergo2025stochastic} is trained to model the distribution of latent trajectories.
A Gaussian sample $\boldsymbol\xi\sim\mathcal{N}(0,I)$ is coupled to each data trajectory. At a randomly sampled flow time $\tau\in[0,1]$, the interpolated latent state is
\begin{equation}
  \mathbf{Z}_\tau = \alpha_\tau \mathbf{Z}_1 + \sigma_\tau \boldsymbol{\xi},
\end{equation}
where Beam3D and the mechanical metamaterials use the generalized variance-preserving (GVP) \cite{ma2024sit} path with $\alpha_\tau = \sin(\frac{1}{2}\pi\tau)$ and $\sigma_\tau = \cos(\frac{1}{2}\pi\tau)$, and Allen-Cahn uses the linear path with $\alpha_\tau = \tau$ and $\sigma_\tau = 1-\tau$.
Moreover, Eq.~\ref{eq:interpolant} in the main text shows the linear path. For a general path, the velocity field of Eq.~\ref{eq:guided-velocity} is $\widehat{v}_{\theta_{\mathrm{flow}}}=\dot\alpha_\tau\widehat{\bfZ}_1+\dot\sigma_\tau\,(\bfZ_\tau-\alpha_\tau\widehat{\bfZ}_1)/\sigma_\tau$, which reduces to $(\widehat{\bfZ}_1-\bfZ_\tau)/(1-\tau)$ for the linear path.
The configured models directly estimate the clean latent trajectory,

\begin{equation}
\widehat{\bfZ}_1
=
f_{\theta_{\mathrm{flow}}}(\bfZ_\tau,\tau,\bfZ^{1:T_{\mathrm{obs}}},\mathbf{c}).
\end{equation}

The predicted latent trajectory is mapped back to the physical signal using the frozen decoder:
\begin{equation}
\widehat{\bfM}^t = \mathcal{D}_{\phi_{\mathrm{dec}}}\left(\widehat{\bfZ}_1^t,\bfX^1\right).
\end{equation}

\subsection{Implementation Details}
\label{app:implementation_details}
\textit{Beam3D}
For Beam3D, each state contains the three-dimensional displacement field of the beam.
The Perceiver encoder represents a state using six latent tokens of dimension $128$, producing a fixed-size representation that is independent of the number of mesh nodes.
The flow model is conditioned on the prescribed scalar tip displacement and observes the initial trajectory frame in the standard configuration.
Because the direction of beam buckling is rotationally symmetric, random azimuthal rotations are applied during training to expose the model to the complete SO(2) orbit of equivalent buckling directions.

\textit{Mechanical Metamaterials}
Each mechanical metamaterial state is represented by a two-dimensional nodal displacement field together with the node positions, node types, and wallpaper-group information.
A Perceiver autoencoder compresses each state into $128$ latent tokens of dimension $128$.
The flow model is conditioned on the prescribed $2\times2$ deformation gradient, represented as a four-dimensional vector.
We group solution branches belonging to the same physical condition during training, and restrict optimal-transport coupling to samples within the same group.
This encourages consistent matching between alternative bifurcation modes without coupling trajectories generated under different physical conditions.

\textit{Allen–Cahn}
Each Allen–Cahn state is a scalar phase field defined on a periodic $64^3$ grid.
A three-dimensional vision transformer divides the field into volumetric patches and encodes them into a structured latent grid.
The flow model is conditioned on the physical parameters ($\epsilon,\mu$).
Temporal and spatial rotary embeddings allow the transformer to distinguish both trajectory time and spatial latent-token position.
The implementation supports periodic translations, sign inversion, axis reflections, and axis permutations as symmetry-preserving transformations of the Allen–Cahn system.

\subsection{Training}
Training is divided into two stages: (i) autoencoder training and (ii) latent flow training.
The first stage involves training the autoencoder to reconstruct individual physical states.
The encoder $\mathcal{E}$ and decoder $\mathcal{D}$ are jointly optimized to reconstruct individual physical states.
Invalid or padded spatial elements are excluded from the reconstruction objective.
The second stage involves training the latent flow model to learn the distribution of latent trajectories.
The pretrained autoencoder is frozen, and only the SiT flow model and physical-parameter conditioner are optimized.
Gaussian noise with the same shape as the latent trajectory is sampled and paired with the data using either independent coupling or a Hungarian optimal-transport assignment based on latent-space squared distance.
When mode groups are available, the assignment is restricted to trajectories belonging to the same physical condition.

For clean-trajectory prediction, the flow time is sampled according to
$\tau\sim\mathcal{U}(\tau_{\min},1-\tau_{\min}),\tau_{\min}=10^{-3}$.

The interpolated latent trajectory $\bfZ_\tau$, flow time $\tau$, control protocol, and observed-frame condition are then passed to the model.
Invalid physical timesteps are replaced by zero before coupling and are excluded from both attention and loss computation.

The standard conditional models observe the first trajectory frame.
The Allen–Cahn transfer configuration uses unconditional generation by setting the number of observed frames to zero.

\paragraph{Optimization}
The second-stage models are optimized with AdamW \citep{loshchilov2017decoupled}.
The standard learning rate is $10^{-4}$, with cosine learning-rate decay for the mechanical metamaterial and standard Allen–Cahn models and a constant learning rate for Beam3D.
Gradient norms are clipped at $1$, and optimizer updates containing non-finite gradients are skipped.
Training is step-based. Checkpoints are selected using the validation loss, with additional checkpoints maintained for generation metrics such as mode coverage.

\paragraph{Data augmentation}
No generic spatial augmentation is applied to the mechanical metamaterial trajectories because their geometric, node-type, and wallpaper-group information is explicitly represented.
Beam3D trajectories are randomly rotated around the longitudinal beam axis, reflecting the continuous rotational symmetry of the buckling problem.
Allen–Cahn augmentation can combine periodic translations, sign inversion, axis reflection, and axis permutation.
Each transformation is applied consistently to all frames of a trajectory to preserve its temporal dynamics.

\subsection{Loss Function}
The first-stage reconstruction model and the second-stage latent flow model are optimized separately.

\paragraph{Reconstruction loss}
For a physical signal $\bfM\in\dR^{N\times C}$ and its reconstruction
$\widehat{\bfM}$, the first-stage reconstruction loss is the mean squared error over nodes:
\begin{equation}
 \mathcal{L}_{\mathrm{rec}} = \frac{1}{N}\sum_{n=1}^{N}\lVert \widehat{\bfm}_n-\bfm_n\rVert_2^2.
\end{equation}
Here, $\bfm_n$ and $\widehat{\bfm}_n$ denote the signal and its reconstruction at node $n$.
Padded or invalid nodes, where present, are excluded from both the sum and the count $N$.
For Metamaterials, $\bfm_n\in\dR^2$ is the nodal displacement.
For Beam3D, $\bfm_n\in\dR^3$ is the three-dimensional nodal displacement.
For Allen-Cahn, $\bfm_n\in\dR$ is the scalar phase-field value.

\textbf{Latent flow loss}
The principal second-stage objective is the frame-masked mean-squared
error between the predicted and target latent trajectories. With
\begin{equation}
\widehat{\bfZ}_1 = f_{\theta_{\mathrm{flow}}}\!\left(\bfZ_\tau,\tau,\bfZ^{1:T_{\mathrm{obs}}},\mathbf{c}\right),
\end{equation}
the loss is
\begin{equation}
\mathcal{L}_{\mathrm{flow}}
=
\frac{
    \displaystyle\sum_{t=1}^{T} r^t
    \left\|
        \widehat{\mathbf{Z}}_1^{\,t}
        - \mathbf{Z}_1^{\,t}
    \right\|_F^2
}{
    L D_z \displaystyle\sum_{t=1}^{T} r^t
}.
\end{equation}
Here, $\tau$ denotes flow time and $t$ indexes trajectory frames.
The tensors $\widehat{\mathbf{Z}}_1^{\,t}, \mathbf{Z}_1^{\,t} \in \mathbb{R}^{L\times D_z}$ are the predicted and target clean latents at frame $t$, respectively.
The indicator $r^t\in\{0,1\}$ specifies whether frame $t$ contributes to the loss.
The denominator normalizes the squared error by the total number of latent entries across contributing frames.

\paragraph{Bifurcation classification loss}
The Metamaterial and Beam3D models optionally include a temporal
attention head that predicts whether each trajectory frame lies
before or after bifurcation. The head is trained using masked
binary cross-entropy:
\begin{equation}
\mathcal{L}_{\mathrm{bif}} =-\frac{\sum_{t=1}^{T} r^t\left[b^t\log p^t+(1-b^t)\log(1-p^t)\right]}{\sum_{t=1}^{T} r^t}.
\end{equation}
Here, $p^t$ is the predicted post-bifurcation probability, $b^t$ is the corresponding ground-truth target, and $\widehat b^t=\mathbbm{1}[p^t\geq 1/2]$ is its binary prediction.

The complete second-stage objective is 
\begin{equation}
\mathcal{L}=\mathcal{L}_{\mathrm{flow}} + \lambda_{\mathrm{bif}}\mathcal{L}_{\mathrm{bif}}.
\end{equation}

The standard Allen–Cahn model does not use the auxiliary classification head and is optimized only using $\mathcal{L}_{\mathrm{flow}}$.





\subsection{Adaptations to particle guidance}
\label{app:guidance}
In their notation, \citet{corso2024particle} add the gradient of a joint cross-particle potential to the reverse-diffusion drift.
They evaluate this interaction on the noisy state $x_\tau$.
We make four adaptations for our latent data-prediction model.

\paragraph{Denoised interaction}
We compute repulsion on the estimate $\widehat{\bfZ}_1^{(k)}$.
It is restricted by the shared predicted mask $\widehat{\bfb}$ in Eq.~\ref{eq:guided-velocity}, which is computed once before guidance begins and held fixed throughout guided sampling.
The original method computes it directly on $x_\tau$.
We scale the resulting displacement by $(\sigma_\tau^2+\delta)^{-1}$ before adding it to the velocity update in Eq.~\ref{eq:guided-velocity}; the constant $\delta>0$ prevents divergence as $\sigma_\tau\to0$.
The interaction strength therefore follows the model's denoising confidence.

\paragraph{Normalized force}
The original force is an unnormalized gradient of a scalar potential.
Our force is a kernel-weighted average of unit pairwise directions.
We divide by the total kernel mass $\sum_j k_{ij}$.
This keeps its scale stable as the number and weight of neighboring particles change.

\paragraph{Self-interaction}
For a translation-invariant kernel, the original self term has zero gradient and vanishes.
Our self term is excluded from the numerator because its direction is undefined.
It remains in the normalizing denominator and therefore damps the interaction.

\paragraph{Frame and time gating}
The original potential acts on the full particle state with a generic coefficient. 
We apply repulsion only to the predicted bifurcation selected by $\widehat{\bfb} = 1$. It is zero elsewhere.
We use the bounded schedule $\gamma(\tau)=\max(\tau^\kappa,\gamma_{\min})$.
The schedule strengthens toward the data endpoint and never switches off.
Here $\kappa>0$ is the schedule exponent and $\gamma_{\min}\in(0,1]$ is its floor; $w_{\mathrm{rep}}$ is the global repulsion weight. No additional per-frame weight is used beyond $\widehat{\bfb}$.

The bandwidth uses a median heuristic over squared, dimension-scaled pairwise distances.
This choice follows the synthetic-experiment code released with \citet{corso2024particle}.
Their paper leaves the kernel and bandwidth domain-specific and does not prescribe a general formula. We provide the latent-flow and guidance-schedule hyperparameters used for the reported models in \cref{tab:latent_hparams}.

\subsection{Hyperparameters}
\begin{table}[ht]
    \centering
    \caption{Hyperparameters of the autoencoder (first stage) used to produce the reported outputs, per dataset.}
    \label{tab:ae_hparams}
    \tiny
    \begin{tabular}{lccc}
        \toprule
        Parameter & Beam3D & Metamaterials & Allen-Cahn \\
        \midrule
        Architecture              & Perceiver AE            & Perceiver AE   & ViT3D + spatial token bottleneck \\
        Params                    & $1.7\times10^6$          & $1.7\times10^6$                  & $6.1\times10^6$ \\
        Input shape                & node states ($N\times C$) & point set (17 groups, 8 node types) & $64{\times}64{\times}64{\times}1$ voxel grid \\
        Patch size                 & --                        & --                                & $8{\times}8{\times}8$ (512 tokens) \\
        Latent shape ($L\times D_z$) & $6\times128$        & $128\times128$                   & $4{\times}4{\times}4$ ($64\times384$) \\
        Encoder blocks (cross/self-attn) & $1\,/\,1$           & $1\,/\,1$                         & ViT depth $6$ (heads 6, embed 192) \\
        Decoder blocks (cross/self-attn) & $1\,/\,1$           & $1\,/\,1$                         & ViT depth $4$ (heads 6, embed 192) \\
        Head dims (enc cross/latent/dec cross/latent) & $16/16/16/16$ & $16/16/16/16$          & -- \\
        Heads (enc cross/latent/dec cross/latent) & $32/16/32/16$ & $32/16/32/16$                & $6$ (enc) $/\,6$ (dec) \\
        Decoder query width        & $160$                     & $160$                             & -- \\
        Activation                 & GELU                      & GELU                              & GELU \\
        Dropout (latent / query)   & $0.0\,/\,0.05$             & $0.0\,/\,0.05$                    & $0.0$ (encoder) \\
        \midrule
        Loss                       & masked MSE                & masked MSE & masked MSE\\
        Optimizer                  & AdamW                     & AdamW                 & AdamW \\
        Learning rate              & $10^{-4}$                  & $10^{-4}$           & $2\times10^{-4}$ \\
        Weight decay               & $0.0$                      & $0.01$                & $0.01$ \\
        Scheduler                  & Cosine (eta\_min $10^{-6}$) & Cosine (eta\_min $1.5\times10^{-5}$) & none \\
        Gradient norm clipping     & $1.0$                      & $1.0$                & $1.0$ \\
        Batch size                 & $128$                      & $128$             & $128$ \\
        Precision                  & bf16-mixed                & bf16-mixed            & bf16-mixed \\
        \bottomrule
    \end{tabular}
    \vspace{2pt}
\end{table}

\begin{table}[ht]
    \centering
    \caption{Hyperparameters of the latent-space (second-stage) flow model used to produce the reported outputs, per dataset.}
    \label{tab:latent_hparams}
    \tiny
    \begin{tabular}{lccc}
        \toprule
        Parameter & Beam3D & Metamaterials & Allen-Cahn \\
        \midrule
        Params (non-AE)            & $7.1\times10^6$            & $17.2\times10^6$             & $31.8\times10^6$ \\
        in\_dim / cond\_dim         & $128\,/\,128$               & $128\,/\,128$                 & $384\,/\,192$ \\
        Hidden size                & $256$                       & $512$                         & $512$ \\
        Depth                      & $5$                          & $4$                           & $8$ \\
        Heads                      & $8$                          & $16$                          & $16$ \\
        MLP ratio                  & $2.0$                        & $2.0$                         & $2.0$ \\
        Bifurcation head           & attention, $\lambda_{\mathrm{bif}}{=}0.1$        & attention, $\lambda_{\mathrm{bif}}{=}0.1$          & none \\
        Conditioner (in/dim/hidden) & $1\,/\,128\,/\,512$         & $4\,/\,128\,/\,256$            & $2\,/\,192\,/\,512$ \\
        Transport path / target    & GVP / data                   & GVP / data                    & linear / data \\
        Sampler                    & DDIM, $\eta{=}0.0$           & DDIM, $\eta{=}0.0$ (eval $\eta{=}1.0$) & DDIM, $\eta{=}0.0$ (eval $\eta{=}1.0$) \\
        \midrule
        Loss                       & flow MSE (frame-masked)     & flow MSE (frame-masked)       & flow MSE (frame-masked) \\
        Optimizer                  & AdamW                        & AdamW                         & AdamW \\
        Learning rate              & $10^{-4}$                    & $10^{-4}$    & $3\times10^{-5}$ \\
        Weight decay               & $0.0$                        & $0.01$                        & $0.01$ \\
        Scheduler                  & none                          & Cosine (eta\_min $10^{-6}$)   & Cosine ($T_{max}{=}500$k, eta\_min $10^{-5}$) \\
        Batch size                 & $64$                          & $16$                          & $8$ \\
        Precision                  & bf16-mixed                   & bf16-mixed                    & fp32 \\
        \midrule
        PG settings ($w_{\mathrm{vel}}$/$w_{\mathrm{rep}}$/$\kappa$) & $1.0\,/\,1.0\,/\,0.0$ & $1.0\,/\,50\,/\,0.25$ & $1.0\,/\,75\,/\,0.025$ \\
        \bottomrule
    \end{tabular}
\end{table}

\subsection{Evaluation Details: Beam3D and Metamaterials}
\label{app:evaluation_beam3D_mm}
We apply a mode-aware evaluation pipeline with task-specific adaptations to the underlying mode set.
For discrete-mode settings, we match generated rollouts to their nearest ground-truth mode using a nearest-mode distance.
For metamaterials, this uses an absolute distance threshold of $0.1$ over the whole trajectory. We compute the mean absolute error over the whole trajectory and sample $16$ trajectories per condition, using $50$ integration steps.
For both Beam3D and Allen-Cahn, the mode space is continuous.
Rollouts for the Beam3D dataset are therefore evaluated by binning the buckling azimuth and matching against the nearest orbit member with a relative-distance rejection criterion.
We reject a generated trajectory if its relative distance to the matched orbit member exceeds $50\%$ of its ground-truth displacement magnitude.
Here we sample $180$ trials with fewer integration steps ($10$ steps, DDIM \citep{song2020denoising}), and compare against GeoTDM ($1000$ steps, DDPM \citep{ho2020denoising}) and STFlow ($10$ steps, Euler integration) using the same number of evaluation samples.

\subsection{Evaluation Details: Allen-Cahn}
\label{app:evaluation_allen_cahn}
The Allen-Cahn dataset dynamics are invariant under periodic translations, sign flips and point-group symmetries of the cubic domain, making pointwise comparison to a single reference rollout not meaningful. We evaluate in two steps:
\begin{enumerate*}[label=(\roman*)]
    \item each generated sample is matched to a ground-truth branch using canonical alignment,
    \item and three different JSD metrics test distributional uniformity over the possible orbit operations.
\end{enumerate*}

\paragraph{Matching} For a generated field $\bfM$, we compute the time-averaged radial power spectrum $S(\bfM)$ as the magnitude of the 3D FFT at each timestep, binned by integer radial frequency and averaged over time.
This value is invariant to translation and the octahedral point group, making it suitable for identifying the best mode independent of orientation and shift.
The matched branch is $m_{\star}^{(k)} = \operatorname*{arg\,max}_{m} \operatorname{cos\_sim}\bigl(S(\bfM^{(k)}),\, S(\mathbf R^{(m)})\bigr)$, where $\mathbf R^{(m)}$ is the selected ground-truth reference for branch $m$.
We count sign-inverted counterparts separately for mode-coverage JSD, but disregard sign for translation and rotoreflection JSD.
Given $m_{\star}^{(k)}$, we recover translation and orientation by an exhaustive search over the 48-element octahedral group $G_{\mathrm{oct}}$ (axis reflections and permutations).
For each $g \in G_{\mathrm{oct}}$, we apply $g$ to the sample's mean field and phase-correlate it against the matched mode's mean field using the normalized cross-power spectrum.
The peak location gives the best periodic shift, and the height gives an alignment score.
We only keep the $g_\star$ with the highest peak, resulting in one match result $(m_{\star}^{(k)},\boldsymbol\psi_{\star}^{(k)},g_{\star}^{(k)})$ per sample, where $\boldsymbol\psi_{\star}^{(k)}\in\mathbb Z_{64}^3$ is the optimal periodic grid shift.
The approach mirrors the alignment step in~\cite{klein2023equivariant}. In contrast to that work, and because the symmetry group is small and finite, the exact minimum is found directly rather than approximated.

\paragraph{MCon-MAE} For each generated rollout $\widehat\bfM^{(k)}$, let $\widehat\bfM^{\mathrm{rel},(k)}$ be its Gauss--Newton relaxation onto an Allen-Cahn PDE-consistent trajectory. We compute
\begin{equation}
    \text{MCon-MAE} = \frac{1}{K}\sum_{k=1}^{K}
    \operatorname{mean}\!\left(\left|\widehat\bfM^{\mathrm{rel},(k)}-\widehat\bfM^{(k)}\right|\right),
\end{equation}
This dataset-specific metric compares each raw generation with its own relaxed projection; no saved reference rollout, matched branch, or branch label enters the calculation. It is therefore distinct from the nearest-reference MCon-MAE in Eq.~\ref{eq:mcon_mae}.

\paragraph{JSD Metrics} Each JSD metric evaluates one component of the matching approach above over $K$ trials.
The results are then aggregated into a categorical distribution, and we compute the Jensen-Shannon divergence against the uniform distribution, as summarized in \cref{tab:jsd_explanation}.

\textbf{Sampling}. To estimate the distribution over the large number of potential solutions, we sample $100$ times for each condition $\mathbf{c}$, using $30$ denoising steps.

\begin{table}[ht]
    \centering
    \caption{Summary of the Allen-Cahn JSD metrics, including their categories and tests.}
    \label{tab:jsd_explanation}
    \small
    \begin{tabular}{l p{4cm} p{4cm}}
        \toprule
        \textbf{Metric} & \textbf{Categories} & \textbf{Tests} \\
        \midrule
        Mode Coverage JSD & mapping to $\mathcal{M}_{\mathbf{c}}$ ground truth branches & Are all solution branches reached equally often?\\
        Translation JSD & grid shifts, coarsened to 8 bins per axis ($8^3 = 512$) & Given one ground truth sample, how often are periodic shifts recovered? \\
        Rotoreflection JSD & 48 octahedral elements & Given one ground truth sample, is the recovered orientation/reflection uniform? \\
        \bottomrule
    \end{tabular}
\end{table}

\section{Results}
\subsection{Quantitative results}
We present a summary of our results for all three datasets in the tables below. We divide the results into two sections: an overview of the JSD and the MCon-MAE results. Section \ref{sec:experiments} explains additional metrics, which we include for each dataset. For the JSD tables, we also incorporate upper and lower bounds as references. The upper bound represents total mode collapse, while the lower bound reflects the expected JSD when sampling from a random uniform distribution. For all tables, we used five different evaluation seeds. 

\textbf{Beam3D}
\begin{table}[H]
    \centering
    \caption{Beam3D prediction performance. Lower MCon-MAE, JSD, and rejection rate indicate better reconstruction accuracy, better agreement with the target mode distribution, and fewer physically invalid rollouts. (mean $\pm$ std over five evaluation seeds)}
    \label{tab:beam3d_results}
    \small
    \begin{subtable}[t]{0.40\linewidth}
        \centering
        \caption{JSD}
        \begin{adjustbox}{max width=\linewidth}
        \begin{tabular}{@{}lc@{}}
        \toprule
        \textbf{Method} & \textbf{JSD} \\
        \midrule
        GeoTDM & $\mathbf{0.036 \pm 0.001}$ \\
        STFlow & $0.074 \pm 0.001$ \\
        \midrule
        \ourmethod~w/o PG & $0.043 \pm 0.003$ \\
        \ourmethod~& $0.041 \pm 0.001$ \\
        \midrule
        Upper bound$^{*}$ & 0.644 \\
        Lower bound$^{**}$ & 0.040 \\
        \bottomrule
        \end{tabular}
        \end{adjustbox}
        \par\smallskip
        {\tiny $^{*}$ assuming total mode collapse, $^{**}$ expected JSD when sampling from a random uniform distribution\par}
    \end{subtable}\hfill%
    \begin{subtable}[t]{0.58\linewidth}
        \centering
        \caption{MCon-MAE and rejection rate}
        \begin{adjustbox}{max width=\linewidth}
        \begin{tabular}{@{}lcc@{}}
        \toprule
        \textbf{Method} & \textbf{MCon-MAE} & \textbf{Rejection} \\
        \midrule
        GeoTDM & $1.040 \pm 0.001$ & $99.2 \pm 0.0$ \\
        STFlow & $0.114 \pm 0.000$ & $0.5 \pm 0.1$ \\
        \midrule
        \ourmethod~w/o PG & $0.018 \pm 0.001$ & $0.1 \pm 0.1$ \\
        \ourmethod~& $\mathbf{0.018 \pm 0.000}$ & $\mathbf{<0.01}$ \\
        \bottomrule
        \end{tabular}
        \end{adjustbox}
    \end{subtable}
\end{table}

\textbf{Metamaterials}
For the dataset \ourmethod~achieves JSD values below the lower-bound reference. This reference represents the expected JSD for a finite set of independent samples from the uniform target distribution, rather than a strict lower bound.
Particle guidance can create dependence between samples that promotes low-discrepancy sampling, leading to more balanced mode frequencies and JSD values below the reference. Full coverage requires a valid sample for every mode; JSD includes all generated samples.
\begin{table}[H]
    \centering
    \caption{Result comparison of \ourmethod~(including w/o particle guidance) against the affine baseline, evaluated with MCon-MAE, JSD, and coverage for each mode count. (mean $\pm$ std over five evaluation seeds)}
    \footnotesize
    \label{tab:results_mm}
    \begin{subtable}[t]{0.47\linewidth}
        \centering
        \caption{JSD}
        \begin{adjustbox}{max width=\linewidth}
        \begin{tabular}{@{}lccc@{}}
        \toprule
        \multirow{2}{*}{\textbf{Method}} & \multicolumn{3}{c}{\textbf{JSD}} \\
        \cmidrule(lr){2-4}
        & $|\mathcal{M}_{\mathbf{c}}|=1$ & $|\mathcal{M}_{\mathbf{c}}|=2$ & $|\mathcal{M}_{\mathbf{c}}|=4$ \\
        \midrule
        Affine Baseline & $0.000$ & $0.216$ & $0.380$ \\
        \ourmethod~w/o PG & $0.000 \pm 0.000$ & $0.050 \pm 0.002$ & $0.087 \pm 0.008$ \\
        \ourmethod~ & $0.000 \pm 0.000$ & $\mathbf{0.013 \pm 0.001}$ & $\mathbf{0.023 \pm 0.002}$ \\
        \midrule
        Upper bound$^{*}$ & 0 & 0.216 & 0.380 \\
        Lower bound$^{**}$ & 0 & 0.008 & 0.027 \\
        \bottomrule
        \end{tabular}
        \end{adjustbox}
        \par\smallskip
        {\tiny $^{*}$ assuming total mode collapse, $^{**}$ expected JSD when sampling from a random uniform distribution\par}
    \end{subtable}\hfill%
    \begin{subtable}[t]{0.51\linewidth}
        \centering
        \caption{MCon-MAE and full coverage}
        \begin{adjustbox}{max width=\linewidth}
        \begin{tabular}{@{}lccc@{}}
        \toprule
        \multirow{2}{*}{\textbf{Method}} & \multirow{2}{*}{\textbf{MCon-MAE}} & \multicolumn{2}{c}{\textbf{Full Coverage \%}} \\
        \cmidrule(lr){3-4}
        & & $|\mathcal{M}_{\mathbf{c}}|=2$ & $|\mathcal{M}_{\mathbf{c}}|=4$ \\
        \midrule
        Affine Baseline & $0.021$ & 0 & 0 \\
        \ourmethod~w/o PG & $\mathbf{0.016 \pm 0.000}$ & $38 \pm 0.9$ & $13 \pm 3.4$ \\
        \ourmethod~& $0.018 \pm 0.000$ & $74 \pm 1.0$ & $48 \pm 3.7$ \\
        \bottomrule
        \end{tabular}
        \end{adjustbox}
    \end{subtable}
    \vspace{-2em}
\end{table}

\textbf{Allen-Cahn}
\begin{table}[H]
    \centering
    \caption{Result comparison of \ourmethod~(including w/o particle guidance), evaluated with relaxed MCon-MAE and three JSD metrics. (mean $\pm$ std over five evaluation seeds)}
    \footnotesize
    \begin{subtable}[t]{0.66\linewidth}
        \centering
        \caption{JSD}
        \begin{adjustbox}{max width=\linewidth}
        \begin{tabular}{@{}lcccc@{}}
        \toprule
        \textbf{Method} & \shortstack{\textbf{Mean JSD}\\\textbf{JSD}} & \shortstack{\textbf{Mode-Coverage}\\\textbf{JSD}} & \shortstack{\textbf{Translation}\\\textbf{JSD}} & \shortstack{\textbf{Rotoreflection}\\\textbf{JSD}} \\
        \midrule
        \ourmethod~w/o PG & $0.217 \pm 0.004$ & $0.131 \pm 0.010$ & $0.446 \pm 0.000$ & $0.076 \pm 0.001$ \\
        \ourmethod~& $\mathbf{0.204 \pm 0.001}$ & $\mathbf{0.089 \pm 0.001}$  & $0.446 \pm 0.000$ & $0.076 \pm 0.002$ \\
        \midrule
        Upper bound$^{*}$ & - & $0.497$ & $0.686$ &$ 0.642$ \\
        Lower bound$^{**}$ & - & $0.009$& $0.446$ & $0.076$ \\
        \bottomrule
        \end{tabular}
        \end{adjustbox}
        \par\smallskip
        {\tiny $^{*}$ assuming total mode collapse, $^{**}$ expected JSD when sampling from a random uniform distribution\par}
    \end{subtable}\hfill%
    \begin{subtable}[t]{0.32\linewidth}
        \centering
        \caption{MCon-MAE}
        \begin{adjustbox}{max width=\linewidth}
        \begin{tabular}{@{}lc@{}}
        \toprule
        \textbf{Method} & \textbf{MCon-MAE} \\
        \midrule
        \ourmethod~w/o PG & $0.0515 \pm 0.001$ \\
        \ourmethod~& $\mathbf{0.0453} \pm 0.000$ \\
        \bottomrule
        \end{tabular}
        \end{adjustbox}
    \end{subtable}
    \label{app:table_ac_results}
\end{table}

\subsection{Ablations}

\paragraph{Bifurcation Head Prediction Accuracy}
To effectively use particle guidance, we predict a bifurcation mask corresponding to the bifurcation time point. We assess the accuracy of the predicted masks on the Beam3D and mechanical metamaterials datasets. The Allen-Cahn method does not use a bifurcation head.
Instead, we set the guidance mask to $\widehat{b}^t=1$ for every valid frame, allowing guidance throughout the trajectory.
We compare the results based on accuracy, precision, recall, F1 score, and balanced accuracy, as shown in \cref{tab:bifurcation_classification}.
\begin{table}[h!]
    \centering
    \caption{Bifurcation-head classification on the complete test splits. Reporting the mean and
    standard deviation over five seeds.}
    \label{tab:bifurcation_classification}
    \small
    \begin{tabular}{lccccc}
        \toprule
        \textbf{Dataset} & \textbf{Accuracy} & \textbf{Precision}
        & \textbf{Recall} & \textbf{F1} & \textbf{Balanced acc.} \\
        \midrule
        \textbf{Beam3D}
        & $0.989 \pm 0.000$ & $0.988 \pm 0.000$
        & $1.000 \pm 0.000$ & $0.994 \pm 0.000$
        & $0.950 \pm 0.001$ \\
        \midrule
        \textbf{Metamaterials}
        & $0.970 \pm 0.000$ & $0.940 \pm 0.002$
        & $0.871 \pm 0.002$ & $0.904 \pm 0.002$
        & $0.930 \pm 0.001$ \\
        \bottomrule
    \end{tabular}
\end{table}

\paragraph{Wall Time}
We analyze the wall time of each tested method on the Beam3D dataset. For the underlying experiments, we used a B300 SXM6 with 288 GB evaluated on one input condition over four random draws. We summarize the resulting times in \cref{tab:wall_clock_time}.

\begin{table}[h!]
    \centering
    \caption{Wall-clock time taken to generate 180 samples of the Beam3D dataset over four random seeds.}
    \label{tab:wall_clock_time}
    \small
    \begin{tabular}{lccc}
    \toprule
    \textbf{Method} & \textbf{Samples} & \textbf{Conditions} & \textbf{Seconds Taken}\\
    \midrule
    \text{GeoTDM}      & 180 & 1 & $7766.36$ \\
    \text{STFlow}      & 180 & 1 & $72.32$ \\
    \text{\ourmethod~w/o PG} & 180 & 1 & $\mathbf{2.94}$ \\
    \text{\ourmethod}        & 180 & 1 & $3.95$ \\
    \bottomrule
    \end{tabular}
\end{table}

\newpage
\subsection{Qualitative Results}
\label{app:qualitative_results}

\paragraph{Beam3D}
\begin{figure}[ht]
    \centering
    \includegraphics[width=\linewidth]{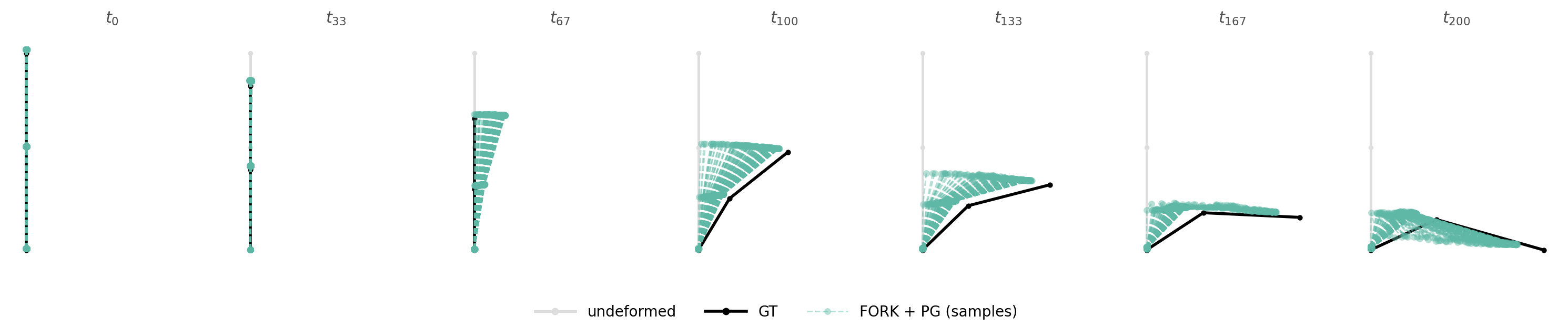}
    \includegraphics[width=\linewidth]{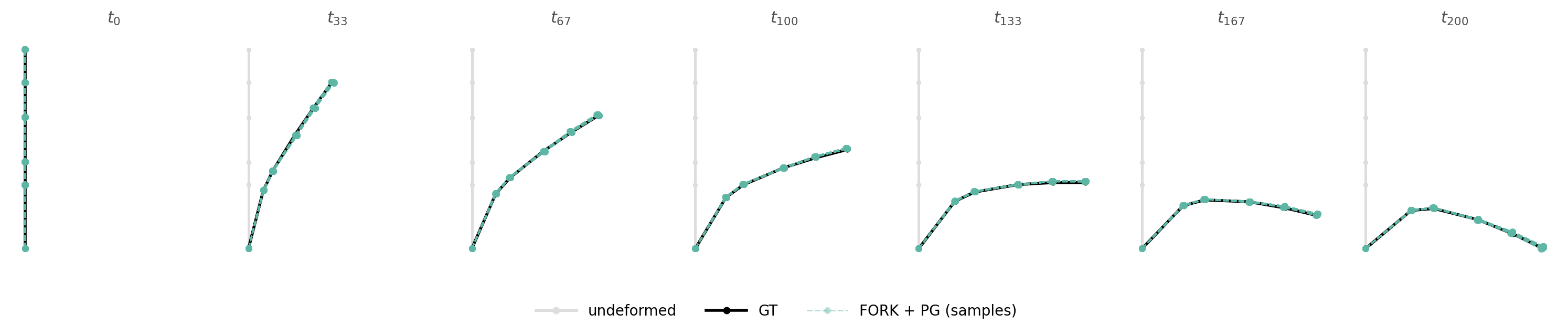}
    \includegraphics[width=\linewidth]{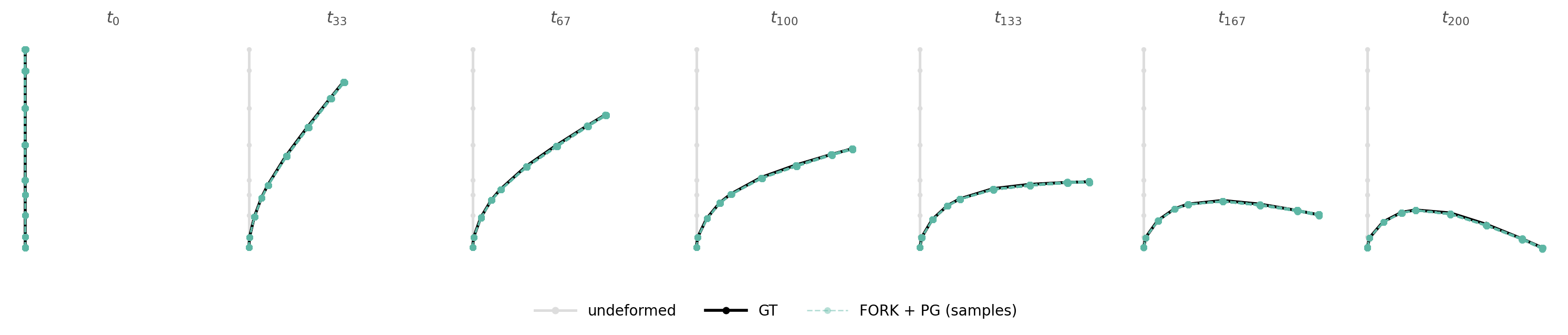}
    \includegraphics[width=\linewidth]{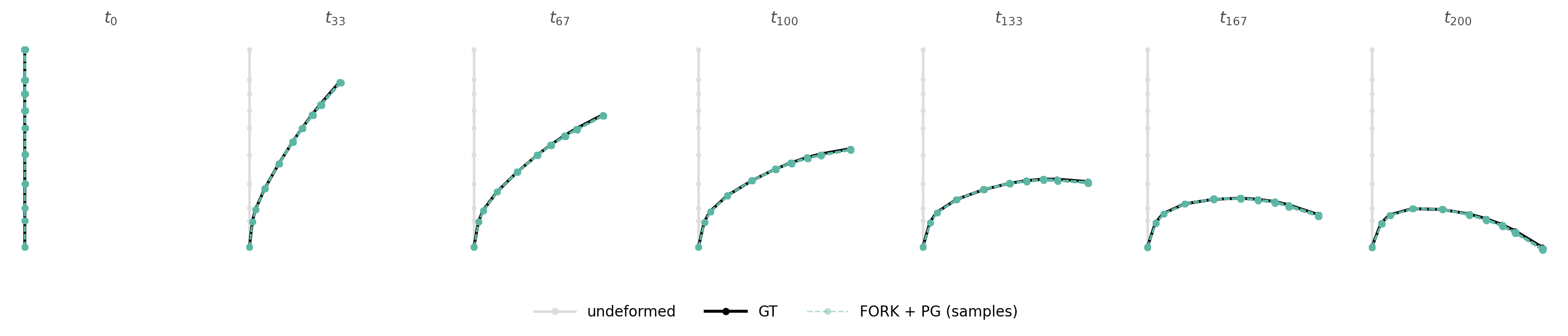}
    \caption{Beam bending process: showing predicted (green) vs actual (black) trajectory evolution for 180 samples. Node count ranging from $3$ to $11$ nodes}
\end{figure}

\textbf{Allen-Cahn}
\begin{figure}[H]
    \centering
    \includegraphics[width=0.9\linewidth]{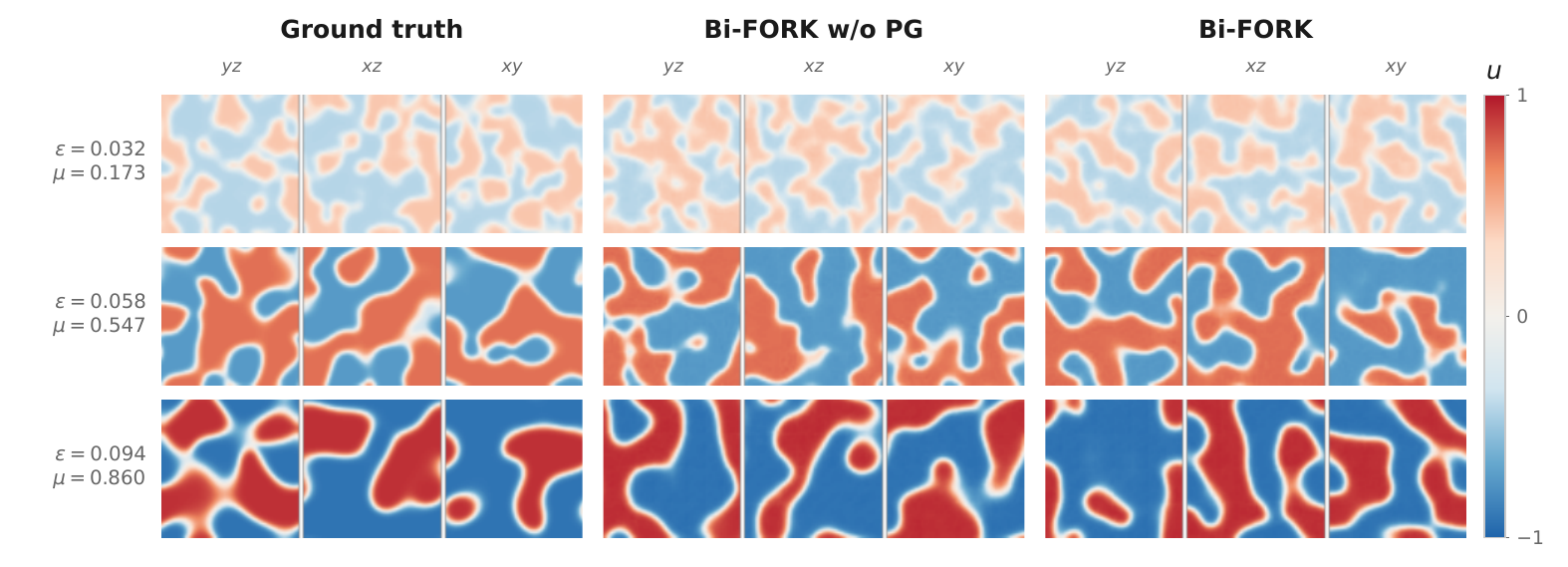}
    \caption{Sample comparison for increasing $\mu$ and $\epsilon$, showcasing the diversity of potential solutions compared to the ground truth.}
\end{figure}

\newpage
\paragraph{Mechanical Metamaterials}
\begin{figure}[H]
    \centering
    \includegraphics[ width=\linewidth]{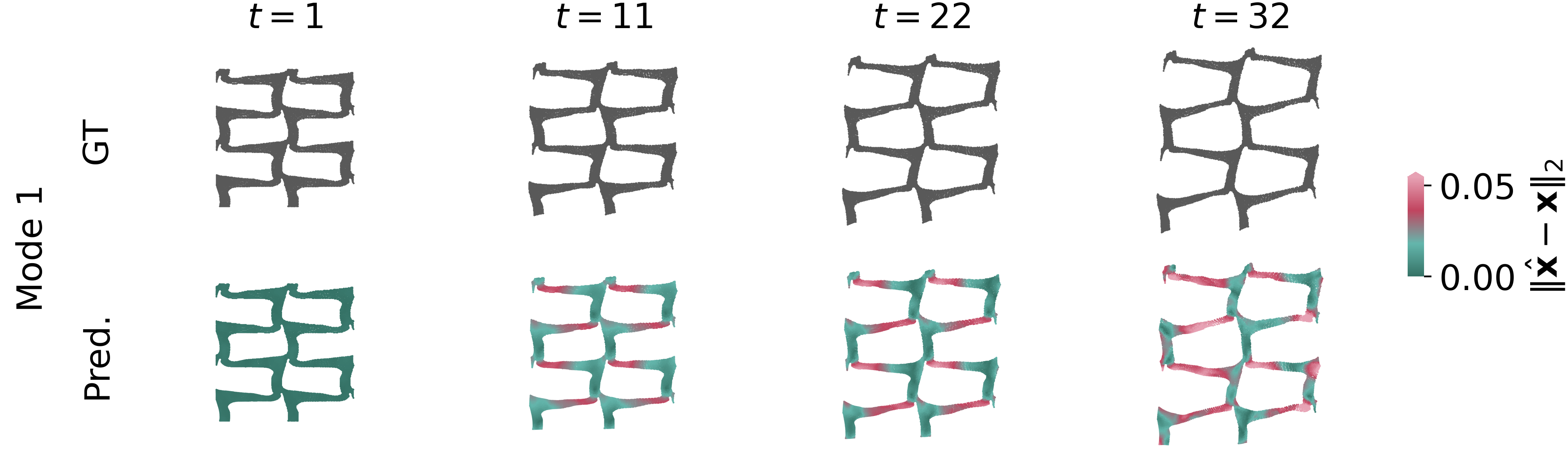}\\[5mm]
    \includegraphics[width=\linewidth]{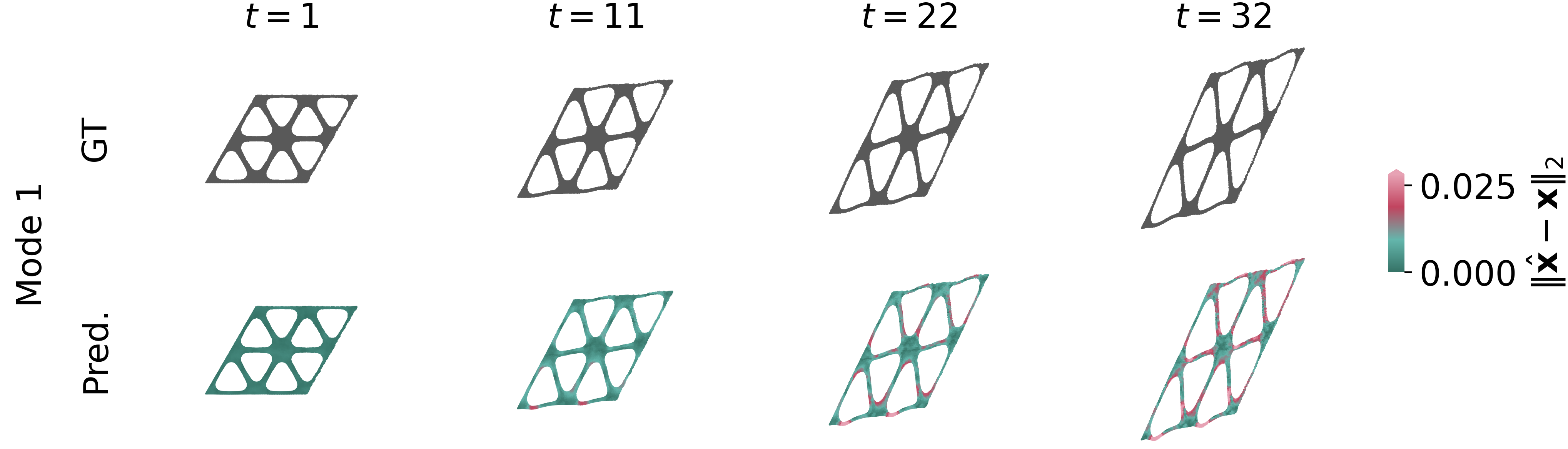}\\[5mm]
    \includegraphics[width=\linewidth]{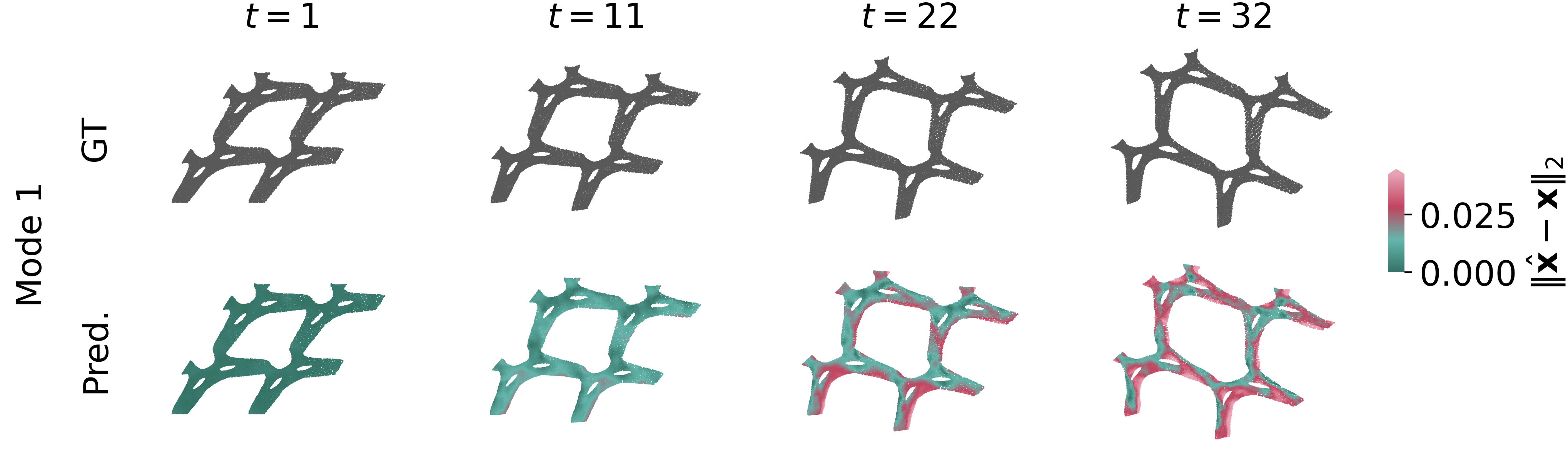}\\[5mm]
    \includegraphics[width=\linewidth]{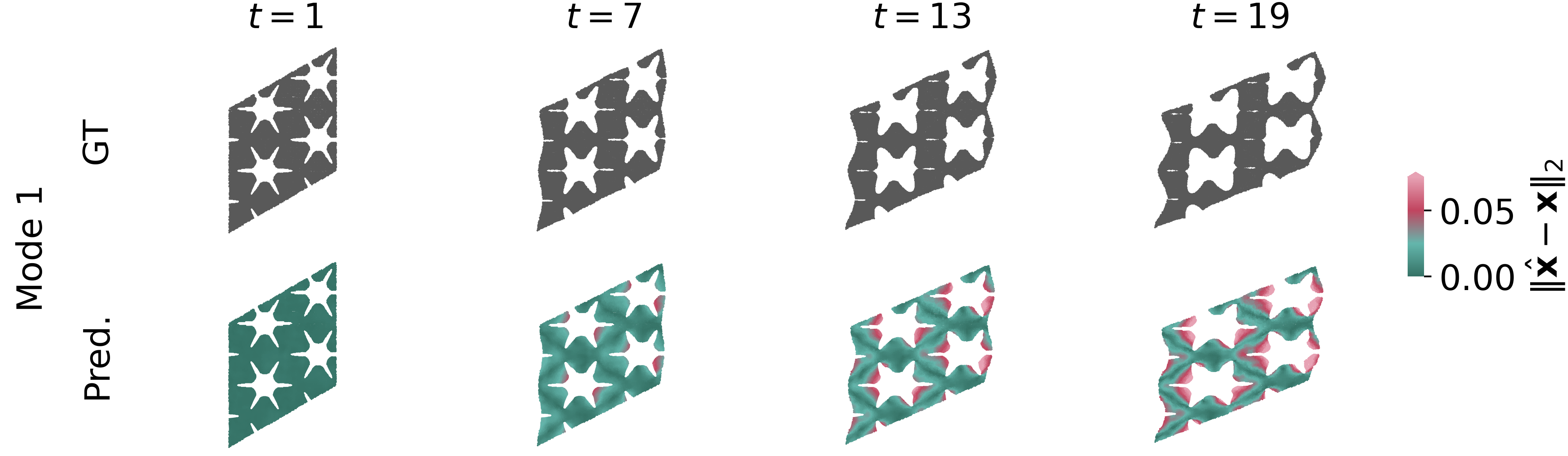}\\[5mm]
    \caption{Visualization of \textbf{one-mode} samples generated by \ourmethod~with $K=16$. The ground truth is compared with the closest predicted mode.}
    \label{fig:m1_mechmaterials}
\end{figure}

\begin{figure}[ht]
    \centering
    \includegraphics[width=\linewidth]{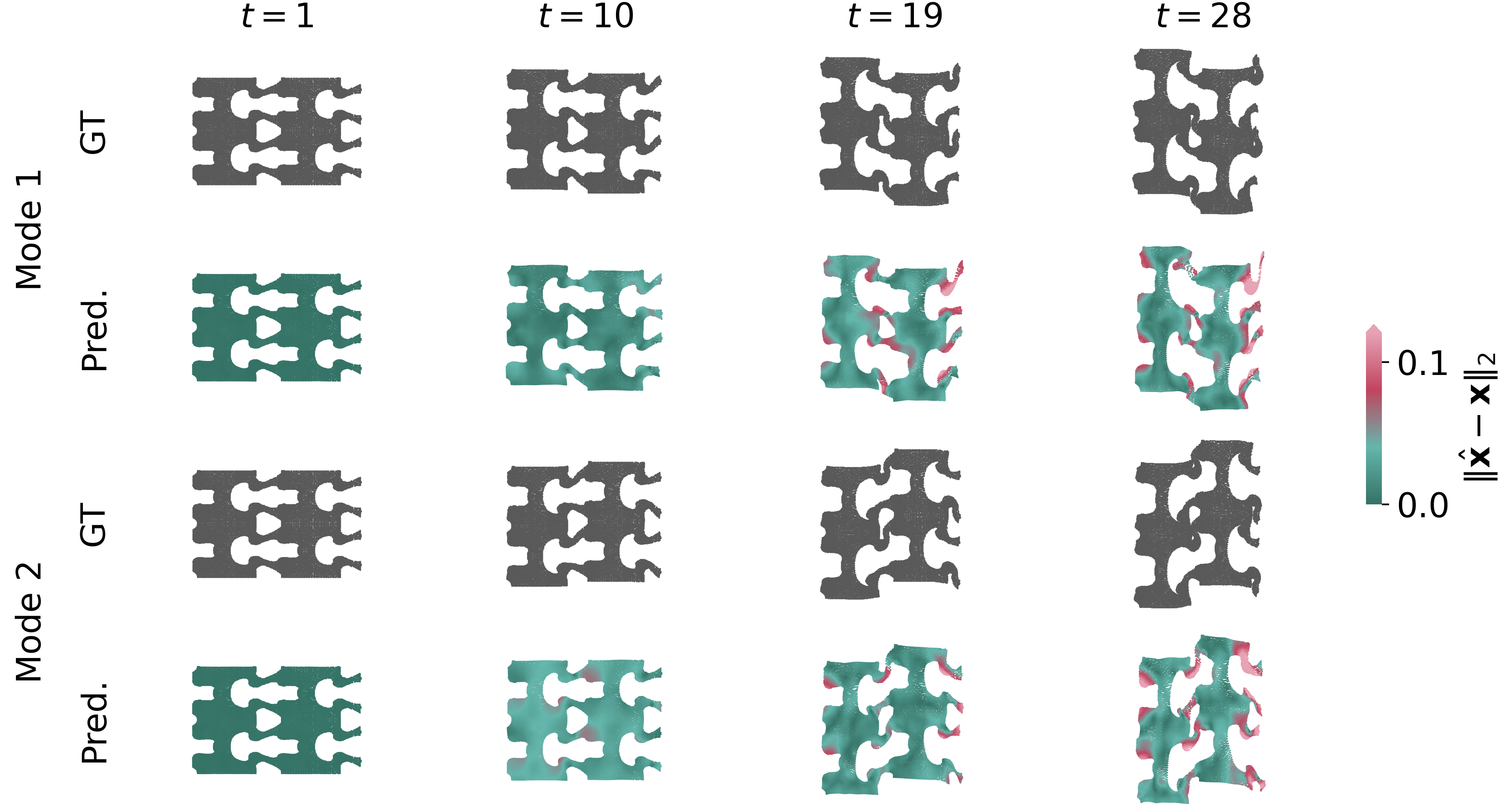} \\[5mm]
    \includegraphics[width=\linewidth]{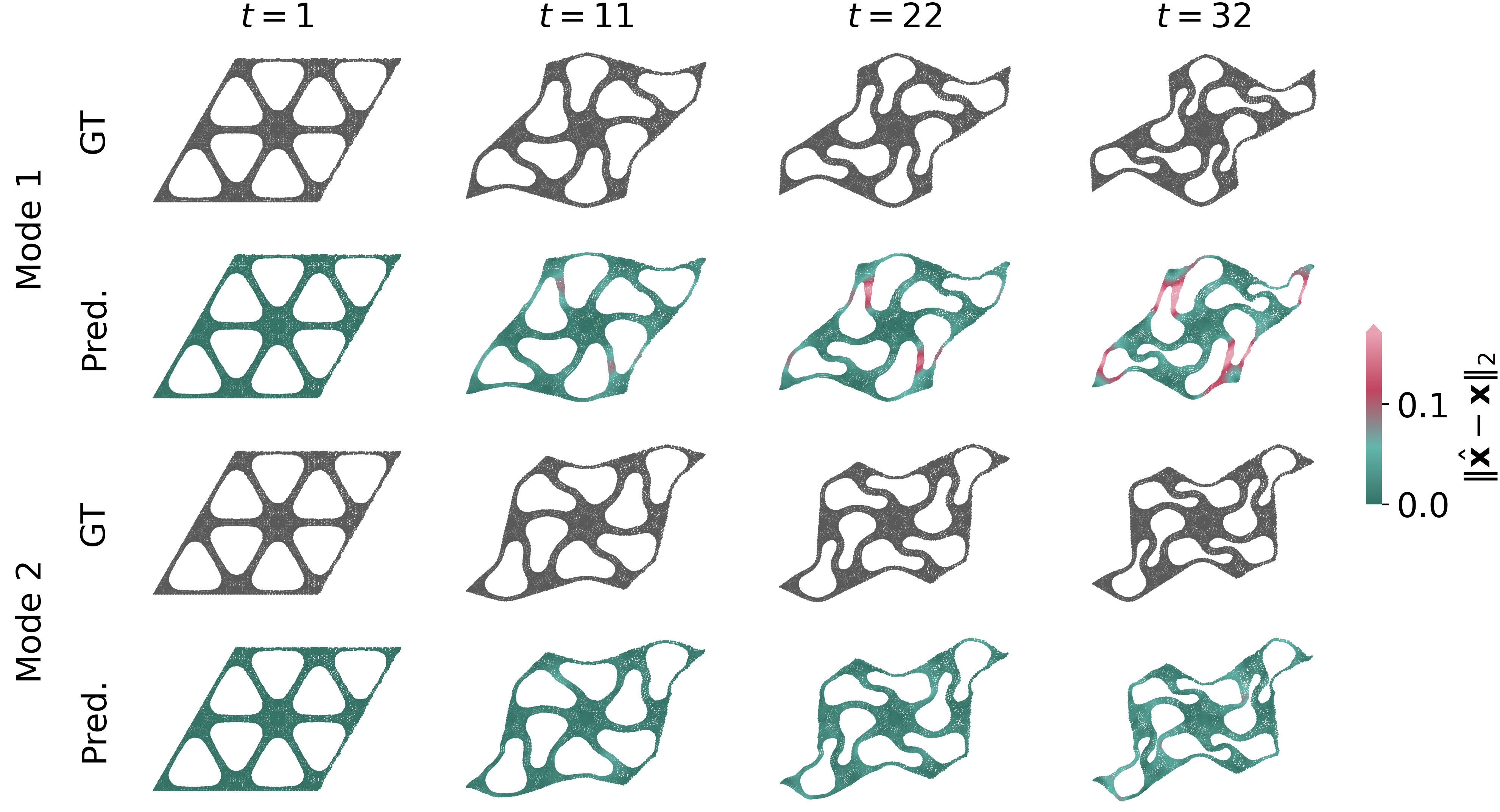} \\[5mm]
    \caption{Visualization of \textbf{two-mode} samples generated by \ourmethod~with $K=16$. The ground truth is compared with the closest predicted mode.}
    \label{fig:m2_mechmaterials}
\end{figure}

\begin{figure}[t]
    \centering
    \includegraphics[trim={0 0 0 0},clip, width=\linewidth]{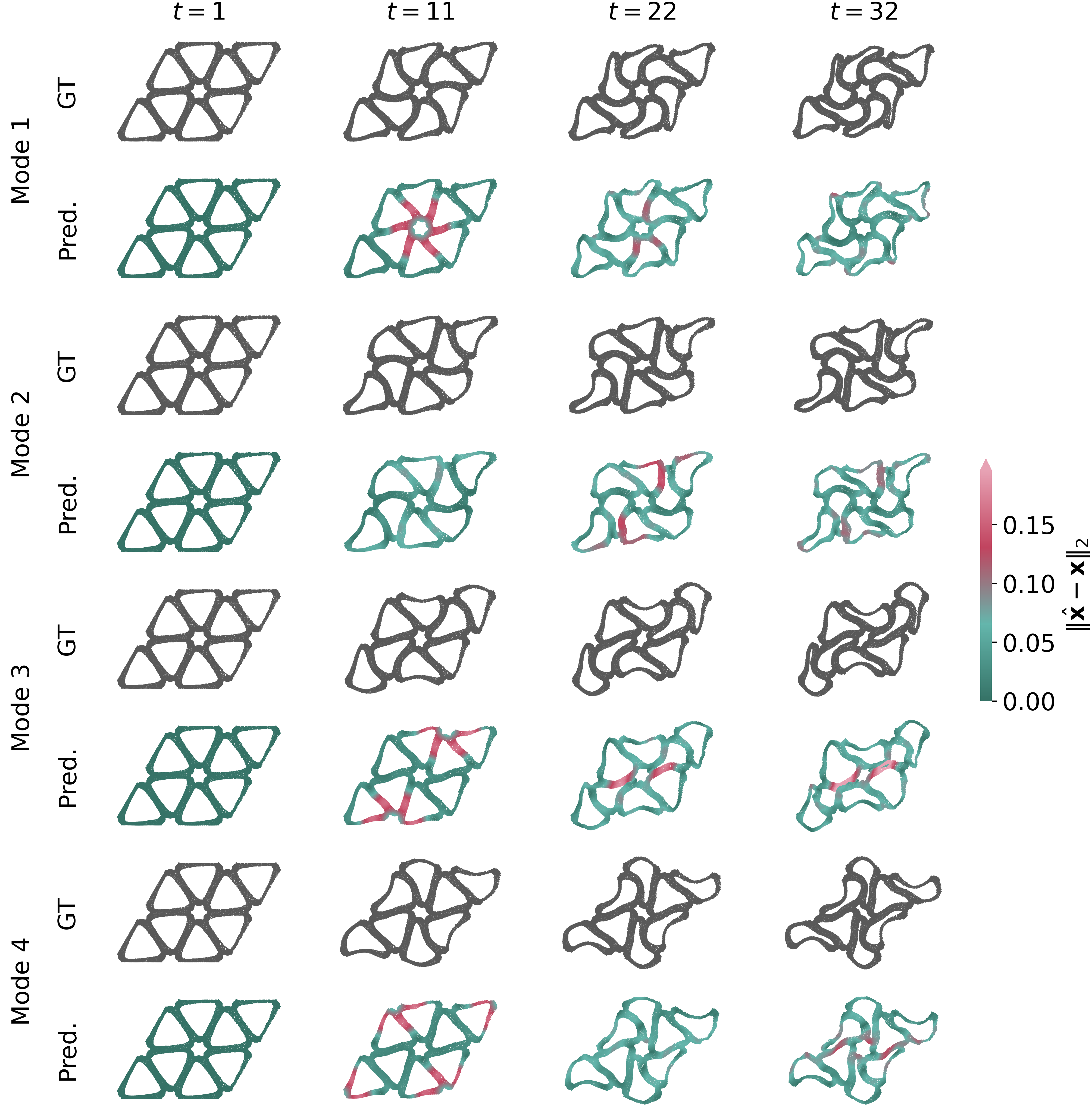}
    \caption{Visualization of a \textbf{four-mode} sample generated by \ourmethod~with $K=16$. The ground truth is compared with the closest predicted mode.}
    \label{fig:m4_mechmaterials}
\end{figure}

\end{document}